\documentclass[journal]{IEEEtran}
\usepackage{graphicx}
\usepackage{subcaption}
\usepackage{multirow}
\usepackage[table]{xcolor}
\usepackage{color}
\usepackage{colortbl}
\usepackage{tabularx}
\usepackage{bm}
\usepackage{booktabs}
\usepackage{url}
\usepackage{stackengine}
\usepackage{float}
\usepackage{verbatim}
\usepackage{array}
\usepackage{algorithm}
\usepackage{algorithmic}
\usepackage[colorlinks,linkcolor=blue,citecolor=cyan]{hyperref}
\usepackage{amsmath,amssymb} 

\begin{document}

\title{Adversarial Shape Learning for Building Extraction in VHR Remote Sensing Images}

\author{Lei~Ding, Hao Tang, Yahui Liu, Yilei Shi,~\IEEEmembership{Member,~IEEE,}~ Xiao Xiang Zhu,~\IEEEmembership{Fellow,~IEEE}~ and Lorenzo~Bruzzone,~\IEEEmembership{Fellow,~IEEE}~
\thanks{L. Ding, Y. Liu, and L. Bruzzone are with the Department of Information Engineering and Computer Science, University of Trento,
38123 Trento, Italy (E-mail: dinglei14@outlook.com, hao.tang@unitn.it, yahui.liu@unitn.it, lorenzo.bruzzone@unitn.it).}
\thanks{H. Tang is with the Department of Information Technology and Electrical Engineering, ETH Zurich,  8092 Zurich, Switzerland. (E-mail: hao.tang@vision.ee.ethz.ch).}
\thanks{Y. Shi is with the Chair of Remote Sensing Technology, Technical University of Munich (TUM), 80333 Munich, Germany (E-mail: yilei.shi@tum.de).}
\thanks{X. Zhu is with the Remote Sensing Technology Institute (IMF), German Aerospace Center (DLR), 82234 Wessling, Germany, and also with the Data Science in Earth Observation (SiPEO, formerly Signal Processing in Earth Observation), Technical University of Munich (TUM), 80333 Munich, Germany (E-mail: xiaoxiang.zhu@dlr.de).}
\thanks{This work is supported by the scholarship from China Scholarship Council under the grant NO.201703170123.}}

\markboth{IEEE Transactions on Image Processing}%
{Shell \MakeLowercase{\textit{et al.}}: Bare Demo of IEEEtran.cls for IEEE Journals}

\maketitle

\begin{abstract}
Building extraction in VHR RSIs remains a challenging task due to occlusion and boundary ambiguity problems. Although conventional convolutional neural networks (CNNs) based methods are capable of exploiting local texture and context information, they fail to capture the shape patterns of buildings, which is a necessary constraint in the human recognition. To address this issue, we propose an adversarial shape learning network (ASLNet) to model the building shape patterns that improve the accuracy of building segmentation. In the proposed ASLNet, we introduce the adversarial learning strategy to explicitly model the shape constraints, as well as a CNN shape regularizer to strengthen the embedding of shape features. To assess the geometric accuracy of building segmentation results, we introduced several object-based quality assessment metrics. Experiments on two open benchmark datasets show that the proposed ASLNet improves both the pixel-based accuracy and the object-based quality measurements by a large margin. The code is available at: \href{https://github.com/ggsDing/ASLNet}{\textit{https://github.com/ggsDing/ASLNet}}.
\end{abstract}

\begin{IEEEkeywords}
Building Extraction, Generative Adversarial Networks (GANs), Image Segmentation, Convolutional Neural Network, Deep Learning, Remote Sensing
\end{IEEEkeywords}


\section{Introduction}\label{sc1}

Shape is an important pattern in the process of visual recognition. Direct modeling of shape patterns in images is challenging since it requires a high-level abstract of the object contours. Among the real-world applications of image recognition techniques, building extraction in very high resolution (VHR) remote sensing images (RSIs) is one of the most interesting and challenging tasks that can benefit greatly from learning the shape patterns. It is important for a wide variety of applications, such as land-cover mapping, urban resources management, detection of illegal constructions, etc.

Conventional building extraction algorithms are based on handcrafted features that often fail to model high-level context information and are highly dependent on parameters. Recently, with the emergence of convolutional neural networks (CNNs) and their applications in semantic segmentation tasks (e.g., vehicle navigation \cite{badrinarayanan2017segnet}, scene parsing~\cite{long2015fcn}, medical image segmentation~\cite{ronneberger2015unet}), a large research interest has been focused on adapting these CNN models to building extraction in VHR RSIs. The CNN-based building extraction methods employ stacked convolution operations to extract the intrinsic content information of images, thus they are more effective in exploiting the context information while they are less sensitive to domain changes. A variety of CNN designs for the semantic segmentation of buildings have been introduced with good results \cite{xu2018building,zhu2020map}.

\begin{figure}[t]
    \centering
    \includegraphics[width=1\linewidth]{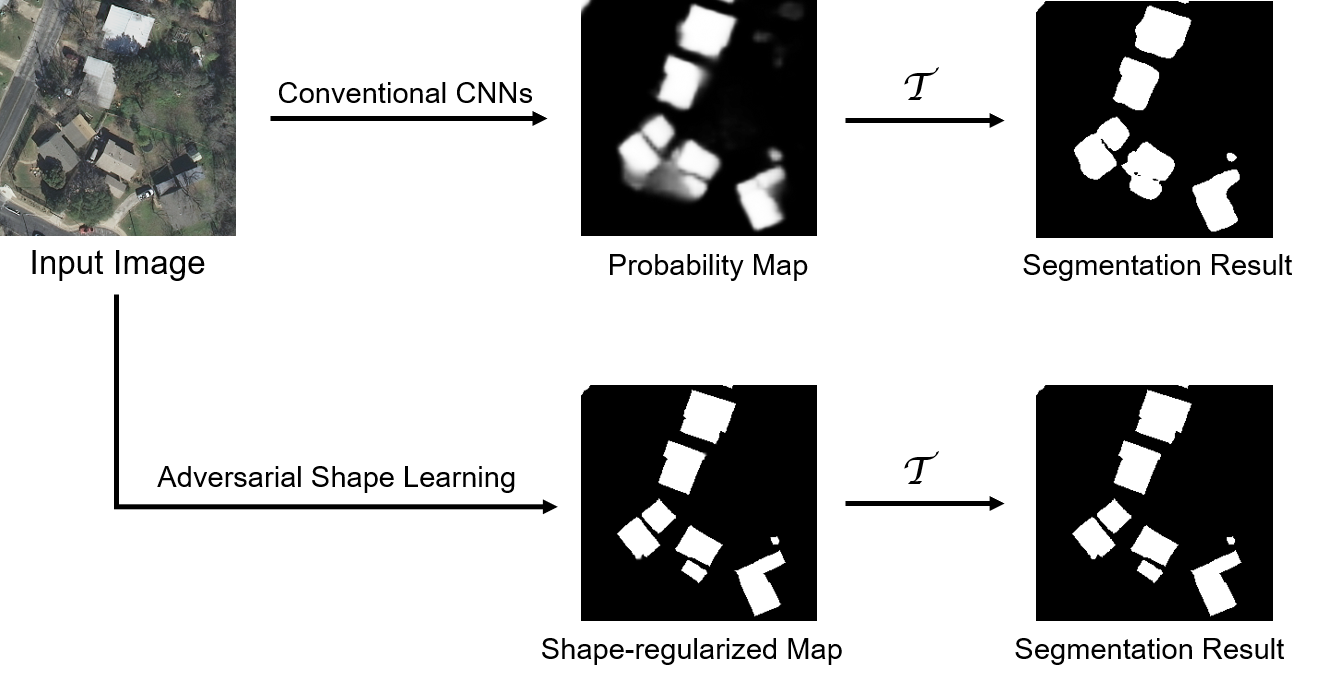}
    \caption{Illustration of the benefits of the proposed shape learning. Conventional CNN models lead to boundary ambiguity problems, whereas the proposed method produces shape-regularized results.}
    \label{FigObjective}
\end{figure}

However, some critical challenges in building extraction remain unsolved even with the use of the recent CNN-based methods. First, occlusions (caused by trees and shadows) and intra-class diversity are common problems in VHR RSIs, which often cause fragmentation and incomplete segmentation. Second, it is common to have boundary ambiguity problems. Due to the effects of shadows and building profiles, an accurate localization of the building boundaries is difficult (especially in the low-contrast areas). Conventional CNN-based methods produce ambiguous probability values in these areas, which often cause rounded or uneven building boundaries after thresholding. Last but not least, the segmentation maps generally suffer from over-segmentation and under-segmentation errors caused by these fragmentation and boundary-adhesion problems. Due to these limitations, post-processing algorithms are often required to optimize the building extraction results \cite{xie2020refined,wei2019toward}.

Another important issue is that previous works on CNN-based building extraction pay more attention to the extraction of texture and context information in RSIs, whereas the explicit modeling of building shapes has rarely been studied. In most cases, buildings in VHR RSIs are compact and rectangular objects with sharp edges and corners. Their rectangularity is very discriminative compared to other ground objects. Learning this shape prior is beneficial for not only inpainting the occluded building parts but also reducing the boundary ambiguities and regularizing the segmentation results. An example is shown in Fig.~\ref{FigObjective} to illustrate the limitations of conventional CNNs and the benefits of the shape modelling.

In this work, we aim to address the previously mentioned issues and to improve the extraction of buildings by introducing an adversarial learning of their shape information.
In greater detail, the main contributions of this work are as follows:

\begin{enumerate}
    \item Proposing an adversarial shape learning network (ASLNet) to learn shape-regularized building extraction results. It includes a shape discriminator to exclude redundant information and focus on modelling the shape information, as well as a shape regularizer to enlarge the receptive fields (RFs) and explicitly model the local shape patterns.
    \item Designing three object-based quality assessment metrics to quantitatively evaluate the geometric properties of the building extraction results. These metrics take into account both the under-segmentation and over-segmentation problems and the shape errors of the predicted building items.
    \item Achieving the state-of-the-art performance on the Inria and Massachusetts building extraction benchmark datasets. Without using sophisticated backbone CNN architectures or post-processing operations, the proposed ASLNet outperforms all the compared literature methods in both pixel-based and object-based metrics.
\end{enumerate}

The remainder of this paper is organized as follows. Section \ref{sc2} introduces the related works on building extraction and adversarial learning. Section \ref{sc3} illustrates the proposed ASLNet. Section \ref{sc4} describes the implementation details and the experimental settings. Section \ref{sc5} presents the results and analyzes the effect of the proposed method. Section \ref{sc6} draws the conclusions of this study.

\section{Related Work}\label{sc2}
\subsection{CNN-based Building Extraction}
Literature work focus on CNN for building extraction can be roughly divided into three types based on the studied perspectives: supervisions, architecture designs and the development of post-processing algorithms. To begin with, while binary ground truth maps are widely used to compute the segmentation losses, several papers have explored the use of other kinds of supervisions. In \cite{yuan2017learning}, the supervision of signed distance map (SDM) is introduced to highlight the difference between building boundaries and inner structures. In \cite{yang2018building} signed distance labels are also introduced but in the form of classification supervision. This SDM has also been used in \cite{shi2020building} as an auxiliary supervision.

Most CNN models for building extraction are variants of the well-known architectures for image classification and semantic segmentation. In \cite{xu2018building}, the ResUNet has been introduced for building extraction from VHR RSIs, which combines ResNet \cite{he2016resnet} with the UNet \cite{ronneberger2015unet} structure. The MFCNN in \cite{xie2020refined} is also a symmetric CNN with ResNet as the feature extractor, whereas it contains more sophisticated designs (such as dilated convolution units and pyramid feature fusion). In \cite{ji2018fully}, a Siamese UNet with two branches is designed to extract buildings from different spatial scales. In \cite{bittner2018building} a hybrid network with multiple sub-nets is introduced to exploit information from the multi-source input data. In \cite{zhu2020map}, the MAPNet is proposed, which is a HRNet-like architecture with multiple feature encoding branches and channel attention designs. In \cite{ma2020building}, the global multi-scale encoder-decoder network (GMEDN) is proposed, which consists of a UNet-like network and a non-local modelling unit.

Since conventional CNN models only produce coarse segmentation results, post-processing operations are often required to obtain detailed results. In \cite{xu2018building}, guided filters are used to optimize the segmented building boundaries and to remove noise. In \cite{wei2019toward} and \cite{zhao2018building}, regularization algorithms are developed to refine the segmentation maps. These algorithms perform object-based analysis on the edges and junction points to generate building-like polygons. In \cite{xie2020refined}, a regularization algorithm is designed based on morphological operations on the rotated segmentation items. In \cite{li2020building}, a graph-based conditional random field (CRF) model is combined with the segmentation network to refine the building boundaries.

\subsection{Adversarial Learning}

\subsubsection{Generative Adversarial Networks (GANs) \cite{goodfellow2014generative}}
GANs typically consist of two important components: a generator and a discriminator. 
The aim of the generator is to generate realistic results from the input data, while the discriminator is used to distinguish between the real data and the generated one. Since the discriminator is also a CNN, it is capable of learning the intrinsic differences between the real and fake data, which can hardly be modeled by human-defined algorithms. Therefore, the GANs have been widely used for a variety of complex tasks in the computer vision field, such as image generation \cite{karras2018style,tang2020xinggan,shaham2019singan,tang2018gesturegan}, semantic segmentation \cite{tsai2018learning,vu2019advent}, object detection \cite{li2017perceptual,wang2017fast}, depth estimation \cite{atapour2018real}, and image/action recognition \cite{tran2019gotta,pan2020adversarial}.

\begin{figure*}[!t]
    \centering
    \includegraphics[width=1\linewidth]{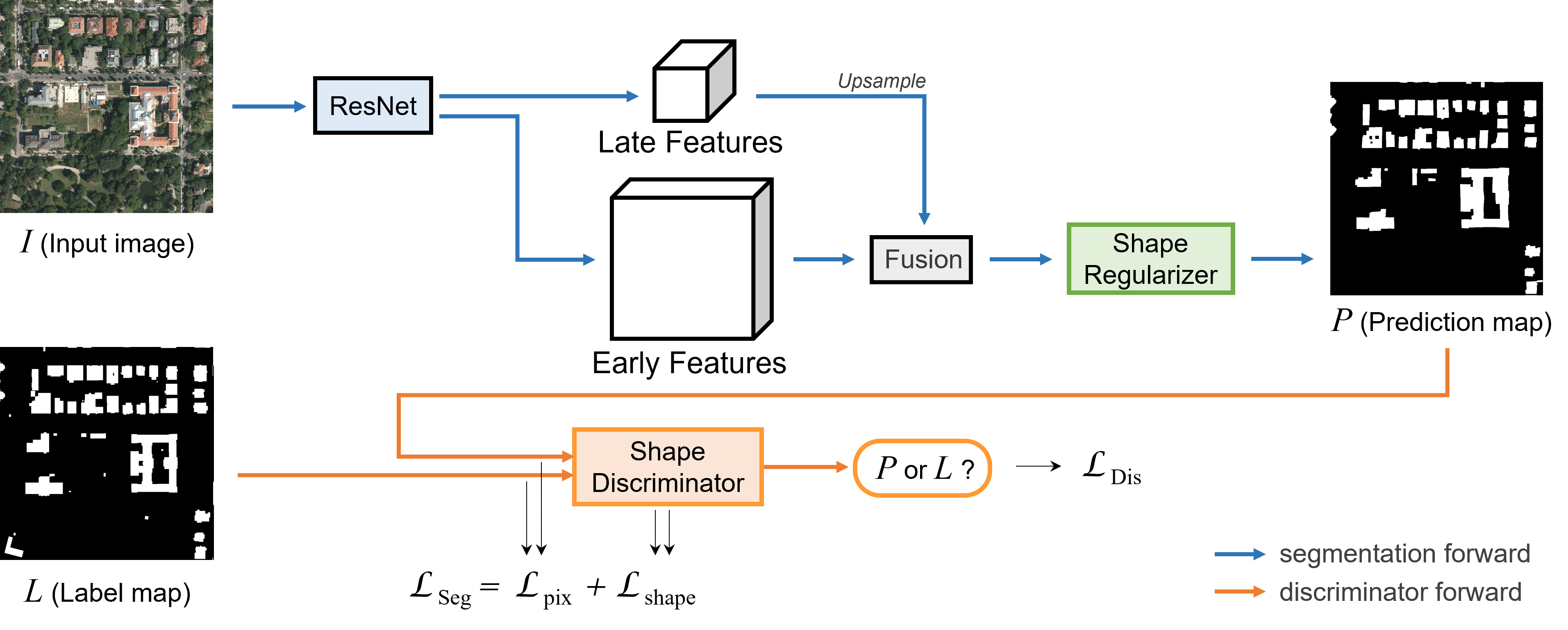}
    \caption{Architecture of the proposed Adversarial Shape Learning Network (ASLNet) for building extraction. We designed an explicit shape regularizer (SR) to model the shape features, and a shape discriminator (SD) to guide the segmentation network. The SD discriminates whether its input is the prediction map ($P$) or the label map ($L$).}
    \label{FigOverview}
\end{figure*}

\subsubsection{Adversarial Learning for Building Extraction}
Several literature works have introduced the adversarial learning strategy for building extraction. The segmentation model can be seen as a generative network, thus the building segmentation results can be learned in an adversarial manner by employing a CNN discriminator. The work in \cite{li2018building} is an early attempt on using the adversarial learning for building extraction. It forwards the masked input RSIs to the discriminator and uses an auto-encoder to reconstruct it.
In \cite{bischke2018overcoming} the GAN has been used to generate synthetic depth maps, thus improving the accuracy of building segmentation. In \cite{abdollahi2020building} the generative adversarial learning is introduced to improve the accuracy of building segmentation by employing a discriminator to distinguish whether the segmentation map is the ground truth (GT) map or the segmentation results. In \cite{pan2019building}, a multi-scale \textit{L$_1$} loss is calculated from the discriminator to train the segmentation network. In \cite{shi2018building}, a conditional Wasserstein GAN with gradient penalty (cwGAN-GP) is proposed for building segmentation, which combines the conditional GAN and Wasserstein GAN.

In general, the literature papers on the use of adversarial learning for building extraction combine the segmentation maps and the RSIs as input data to the discriminator, whereas they do not exploit the shape of segmented items.

\subsection{CNN-based Shape Modelling}
There is a limited number of papers on CNN-based modelling of 2D shapes. To begin with, the work in \cite{atabay2016binary} shows that CNNs can recognize shapes in binary images with high accuracy. In \cite{ravishankar2017learning}, the modelling of shape information is studied for the segmentation of kidneys from ultrasound scan images. In this work, a CNN auto-encoder is introduced to regularize the CNN output, which is pre-trained to recover the intact shape from randomly corrupted shapes. The shape regularization network is trained by three loss terms that measure the distance between the input segmentation map, regularized segmentation map, and the ideal segmentation map. In \cite{takikawa2019gated}, a gated shape CNN is proposed for the semantic segmentation. It contains an explicit shape stream that deals with the object boundary information.

Several works use binary mask features to preserve and model the shape information. In \cite{kuo2019shapemask}, the shape priors are modeled to improve the instance segmentation. The label masks are cluttered to generate class-wise shape priors. These priors are then weighted by a learnt vector of parameters to estimate the coarse instance region. In \cite{ding2019semantic}, a shape-variant convolution is proposed for the semantic segmentation. It uses a novel paired convolution to learn context-dependent masks to limit the receptive fields (RFs) on interested image regions. In \cite{liang2020polytransform}, the modeling of object contour polygons is studied for the instance segmentation. The polygons are first generated with a segmentation CNN and then transformed in a transformer network to fit to the object contours.

To the best of our knowledge, there is no existing work that explicitly models shape constraints for the segmentation of remote sensing images.
\section{Adversarial Shape Learning Network}\label{sc3}

Typical CNN models \cite{xu2018building,xie2020refined} for building segmentation exploit only the local texture and context information, thus the fragmentation and boundary ambiguity problems remain unsolved. Since buildings in VHR RSIs usually have clear shape patterns, it is meaningful to use the shape constraints to alleviate these problems. To this end, we propose the adversarial shape learning network (ASLNet) to explicitly model these shape constraints. In this section, we describe in detail the architecture, loss functions, and the CNN modules of our ASLNet.

\subsection{Network Architecture}

Fig.~\ref{FigOverview} illustrates the architecture of the proposed ASLNet for building extraction, which consists of a segmentation network and a discriminator network. The segmentation network itself is capable of segmenting buildings, while the discriminator is employed to guide the training of the segmentation network. The segmentation network follows the classic encoder-decoder structure in literature papers~\cite{ronneberger2015unet,ding2020lanet,chen2018deeplabv3+}. The encoder network contains down-sampling operations to extract high-level semantic features from image local patches, whereas the decoder network recovers the spatial resolution of encoded features. 
The choice of the encoder network is not the focus of this work, thus we simply adopt the ResNet \cite{he2016resnet} as the feature encoder. 
It has been widely used for feature extraction in building segmentation \cite{liu2019building}, road segmentation \cite{ding2020diresnet}, and other semantic segmentation related tasks \cite{ding2020twostage}. The selected ResNet version is ResNet34, which can be replaced by other versions based on the complexity of the dataset.

Apart from the output features from the late layers of the ResNet (with 1/8 of the original GSD), the early features (with 1/4 of the original GSD) are also employed in the decoder to learn finer spatial details. This is a commonly adopted design in segmentation networks~\cite{chen2018deeplabv3+,ding2020lanet}. This ResNet with encoder-decoder structure is a modified version of FCN~\cite{long2015fcn}, denoted as ED-FCN. Compared with the plain FCN, the ED-FCN models the spatial features at a finer resolution, which is essential for the segmentation of VHR RSIs. It is therefore set as the baseline method of our segmentation network. Building on top of the ED-FCN, we further designed a shape regularizer at the end of the segmentation network in the proposed ASLNet to produce shape-refined outputs.

\begin{figure}[t]
    \centering
    \includegraphics[width=1\linewidth]{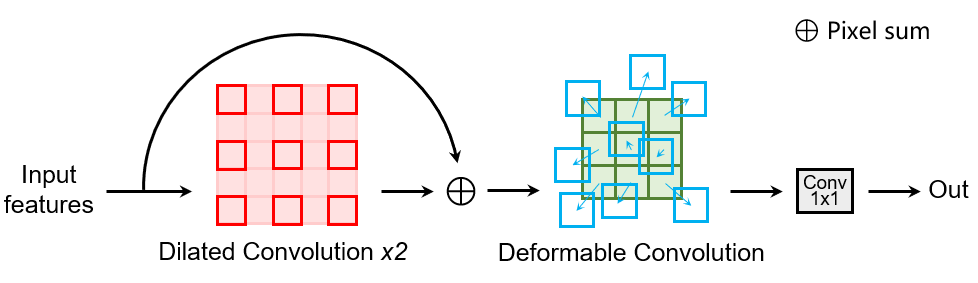}
    \caption{The designed shape regularizer. Dilated convolutions and deformable convolutions are employed to enlarge the RFs and learn the shape features.}
    \label{FigDiscriminator}
\end{figure}

\subsection{Shape Regularizer}

Although using a simple ResNet as the segmentation network is feasible for the adversarial shape learning, it is beneficial to model the shape features at finer spatial scales. Therefore, we design an explicit shape regularizer in the decoder of the segmentation network to enable a better adaptation to the shape constraints (see Fig.~\ref{FigDiscriminator}). The shape regularizer is placed at the spatial scale of 1/4 of the GSD, which operates on the fused multi-scale features in the ED-FCN. This spatial resolution for shape modeling is adopted following the practice in \cite{chen2018deeplabv3+} and \cite{ding2020lanet}, which is a balance between accuracy and computational costs. At this spatial scale, a conventional $3\times 3$ convolutional kernel has the RF of around $12 \times 12$ pixels, which is too small for modelling the local shape patterns. Therefore, we introduce the dilated convolution (DC) and deformable convolution (DFC) \cite{zhu2019deformable} layers to enlarge the RFs and to learn shape-sensitive transformations.

Both the DC and DFC are based on the idea of enlarging the coverage of convolutional kernels. Let us consider a convolutional operation for pixel $x(r,c)$ as:
\begin{equation}
    U(r,c)=\sum_{i,j}x_{r+i,c+j}\cdot k_{i,j},
\end{equation}
where $k_{i,j}$ denotes the kernel weight. In a standard $3 \times 3$ convolution, $i,j \in \{-1,0,1\}$. However, in a $3 \times 3$ DC, $i,j \in \{-r,0,r\}$ where $r$ is the dilation rate. In the designed SR, two $3 \times 3$ DCs are connected in a residual manner as in \cite{he2016resnet}, composing a dilated residual unit. The residual branch allows the unit to gather information in different spatial ranges. In this way, the RF is enlarged to over $36 \times 36$ pixels.

A DFC is further employed to exploit the shape information, defined as:
\begin{equation}
    U_{df}(r,c)=\sum_{i,j}x_{r+i+u(r,c),c+j+v(r,c)}\cdot k_{i,j},
\end{equation}
where $u(r,c)$ and $v(r,c)$ are position parameters learned by the additional convolutions, as follow:
\begin{equation}
    u(r,c)=\sum_{i,j}x_{r+i,c+j}\cdot k^{'}_{i,j}, v(r,c)=\sum_{i,j}x_{r+i,c+j}\cdot k^{''}_{i,j}.
\end{equation}
The DFC is placed at the end of the convolutional module (SR) as in \cite{zhu2019deformable}. This enables the SR to perceive and adapt to the local shape patterns. Finally, a $1 \times 1$ convolution is followed to project the learned features into a segmentation map.

\subsection{Shape Discriminator}

A CNN model (even equipped with the SR) trained by the standard pixel-wise losses is not shape-aware, since each pixel is considered separately. To address this limit, we introduce a shape discriminator (SD) to drive the model to learn shape patterns.
Although several literature works have introduced the adversarial learning for building extraction, most of them combine CNN outputs and input RSIs to train the discriminators~\cite{shi2018building,bischke2018overcoming,abdollahi2020building,pan2019building}. Under this condition, the discriminators are unlikely to learn the shape information, since they are affected by the redundant information in input RSIs. In the proposed ASLNet, the discriminator focuses only on the shape features, thus we exclude the use of input RSIs.

Training a shape discriminator with only binary inputs is challenging. Let $I$ denote an input image, $P$ be its corresponding prediction output and $L$ be the ground truth map. Since in $I$ there are usually mixed pixels (due to the sensor resolution) and discontinuities in objects representations (due to occlusions and low illumination conditions), it is common to have fuzzy areas in especially the building contours in the normalized prediction map $\sigma(P)$, where $\sigma$ is the Sigmoid function. However, in $L$ the human-annotated building contours have 'hard' edges, i.e. $L\in\{0, 1\}$. Mathematically, let $\sigma(P)\in[0,1]$ be a smooth/fuzzy representation of the contours. This difference between $\sigma(P)$ and $L$ can be easily captured by the discriminator and causes failure to the shape modelling. In some literature works \cite{li2018building} a thresholding (or argmax) function $\mathcal{T}$ is employed to binarize $\sigma(P)$ as:
\begin{equation}
    R=\mathcal{T}[\sigma(P)],
    \label{Formula.thresholding}
\end{equation}
where $R$ is the binary segmentation map. Although the obtained $R\in\{0,1\}$, the $\mathcal{T}$ is non-differential in most cases, thus training the segmentation network with $R$ and $L$ will lead to zero-gradient problems.

\begin{figure}[t]
    \centering
    \includegraphics[width=8cm]{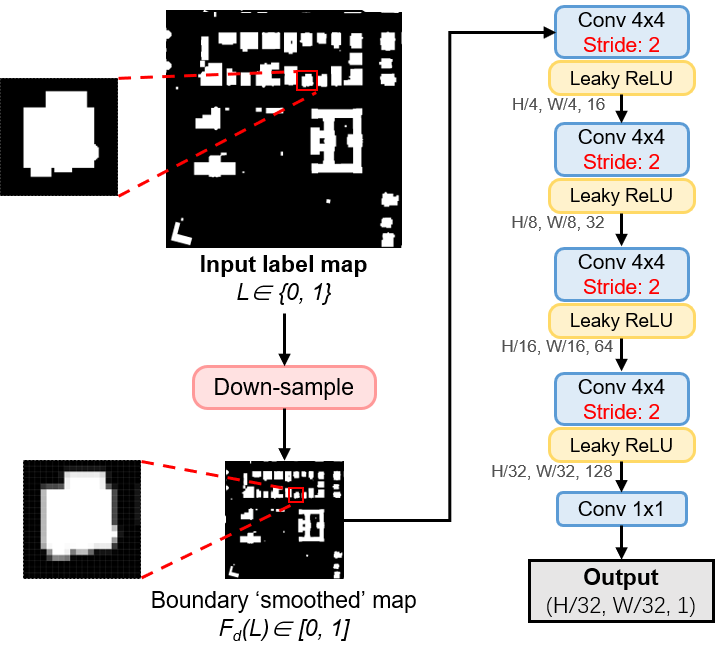}
    \caption{The designed shape discriminator. The input maps are down-scaled to exclude the impact of `hard' building boundaries in reference maps.}
    \label{FigRegularizer}
\end{figure}

In the designed shape discriminator we managed to eliminate this boundary difference and model only the shape information by adding a down-sampling operation $F_d$ in the discriminator $\mathcal{D}$. Fig.~\ref{FigRegularizer} illustrates the designed shape discriminator. After applying $F_d$, the building boundaries in $F_d(L)$ are 'softened' ($F_d(L)\in[0,1]$) and the boundary difference between $F_d(\sigma(P))$ and $F_d(L)$ is excluded. Specifically, four layers of strided convolution and activation functions are then employed to reduce the spatial size of feature maps and learn the local discriminative shape information. The output results are related to 1/32 of the original GSD.

The discriminator is trained with the Binary Cross Entropy (BCE) loss function. It is calculated as:

\begin{equation}\label{Fml.LossD}
\begin{aligned}
\mathcal{L}_{Dis} = & \mathbb{E}_{L\sim{p_{\rm data}}(L)}[\log \mathcal{D}(L)] \\ 
+ & \mathbb{E}_{P\sim{p_{data}(P)}}[\log (1 - \mathcal{D}(\sigma(P)))] \\
= & -y \log(p)-(1-y)\log(1-p),
\end{aligned}
\end{equation}
where $\mathbb{E}$ is the expected value for different types of input samples, $y$ is the encoded signal that depending on the input map to the discriminator can be $L$ or $\sigma(P)$ (`1' and `0', respectively), and $p$ is the output of the discriminator. We employ the Mean Squared Error (MSE) loss function to calculate the $\mathcal{L}_{Shape}$ as:
\begin{equation}\label{Fml.LossG}
    \mathcal{L}_{Shape} = \{\mathcal{D}(L) - \mathcal{D}[\sigma(P)]\}^2,
\end{equation}
where $\mathcal{D}$ is the shape discriminator. In this way, the $\mathcal{L}_{Shape}$ is related to the $L$, thus the segmentation network is constrained by the ground truth conditions.

\subsection{Optimization Objective of ASLNet}

Let $\mathcal{L}_{Seg}$ be the loss function for the CNN-based segmentation of buildings. In conventional CNNs, $\mathcal{L}_{Seg}$ is only related to the pixel-wise accuracy, which does not consider the image context. 
In order to model the shape of objects with CNNs, it is essential to define a shape-based loss function $\mathcal{L}_{Shape}$. Previous works on shape analysis are often object-based \cite{ye2018review,lizarazo2014accuracy}. They include non-differential operations to calculate the shape measures, which are difficult to be incorporated into CNNs. Although there are also literature papers that use CNNs to regularize the shape of predictions \cite{ravishankar2017learning}, pre-training is often required and the regularization is limited to certain functions (e.g., inpainting of object contours). Since CNNs themselves can be trained to discriminate different shapes, we introduce the idea of adversarial learning to learn the $\mathcal{L}_{Shape}$ to guide the segmentation network.

\begin{equation}\label{Fml.LossSeg}
\begin{aligned}
\mathcal{L}_{Seg} = & \alpha \cdot \mathcal{L}_{Pix} + \beta \cdot \mathcal{L}_{Shape}  \\
= & \alpha \cdot [L - \sigma(P)]^2 + \beta \cdot \{\mathcal{D}(L) - \mathcal{D}[\sigma(P)]\}^2,
\end{aligned}
\end{equation}
where $\mathcal{L}_{Pix} = [L - \sigma(P)]^2$ is the supervised pixel-based reconstruction loss, $\alpha$ and $\beta$ are two weighting parameters. The first term in this formula drives the segmentation network to segment pixel-based $P$ in order to fit $L$, while the second term strengthens the local shape similarities between $P$ and~$L$.

\section{Design of Experiments}\label{sc4}
In this section, we describe the experimental dataset, the implementation details, and the considered evaluation metrics.

\subsection{Dataset Descriptions}
We conduct building extraction experiments on two VHR RSI datasets, i.e.,
the Inria dataset \cite{maggiori2017can} and the Massachusetts Building dataset \cite{MnihThesis}. These are two of the most widely studied building extraction datasets in the literature \cite{xie2020refined,li2018building,ma2020building,liu2019building}.

\subsubsection{Inria Dataset \cite{maggiori2017can}}
This is an aerial dataset with the GSD of 0.3 $m$ per pixel, covering 810 $km^2$. Each image has 5,000 $\times$ 5,000 pixels. There is a total of 360 images in this dataset, among which 180 are provided with the ground truth labels. These 180 images were collected in five different cities: Austin (U.S.), Chicago (U.S.), Kitsap (U.S.), Tyrol (Austria), and Vienna (Austria). Following the practice in \cite{xie2020refined,ma2020building}, we use the first 5 images in each city for testing and the rest 31 images for training.

\subsubsection{Massachusetts (MAS) Building Dataset \cite{MnihThesis}}
This is an aerial dataset collected on the Boston area. It has a GSD of 1.2 $m$ per pixel, covering around 340 $km^2$. The imaged regions include urban and suburban scenes where there are buildings with different sizes. This dataset consists of a training set with 137 images, a validation set with 4 images, and a test set with 10 images. Each image has 1,500 $\times$ 1,500 pixels.

\subsection{Implementation Details}
The experiments were conducted on a workstation with 32 GB RAM and a NVIDIA Quadro P6000 GPU (23GB). Since it is impossible to train directly the large RSIs, they are randomly cropped into 512 $\times$ 512 patch images during the training process. The performed data preprocessing and augmentation operations include data normalization, random cropping, and image flipping. The training batch size is set to 8 and the number of training epochs is 50. The validation and test sets are evaluated on the original size RSIs to avoid the impact of cropping parameters. The hyper-parameters $\alpha, \beta$ in the Eq.~\eqref{Fml.LossSeg} are set to 5.0, 1.0, respectively. The choice of hyper-parameters is discussed in Section \ref{sc5.ablation}.

\subsection{Evaluation Metrics}

\subsubsection{Pixel-based Evaluation Metrics}
We adopt several commonly used evaluation metrics in building extraction \cite{xie2020refined,shi2020building} and other binary segmentation tasks \cite{ding2020diresnet} to assess the accuracy of the results. These metrics are based on statistical analysis of the classified pixels, including: overall accuracy ($OA$), Precision ($P$), Recall ($R$), F1 score, and mean Intersection over Union (IoU). The calculations are:
\begin{equation}
    P=\frac{TP}{TP+FP},\quad R=\frac{TP}{TP+FN},
\label{formular_PR}
\end{equation}
\begin{equation}
    F1=2\times\frac{P \times R}{P+R},\quad OA=\frac{TP+TN}{TP+FP+TN+FN},
\label{formular_F1OA}
\end{equation}
\begin{equation}
    IoU=\frac{TP}{TP+FP+FN},
\label{formular_IoU}
\end{equation}
where $TP$, $FP$, $TN$, and $FN$ represent true positive, false positive, true negative, and false negative, respectively.

\subsubsection{Object-based Evaluation Metrics}

Although the pixel-based evaluation metrics present the overall classification accuracy of the results, they fail to consider the thematic and geometrical properties of the segmented units \cite{ye2018review}. To overcome this limitation, we designed three object-based evaluation metrics, including the matching rate ($MR$), the curvature error ($E_{curv}$), and the shape error ($E_{shape}$). These metrics are variants of the literature works \cite{persello2009novel,lizarazo2014accuracy} to adapt to the assessment of building extraction results.

\begin{figure}[t]
\centering
    \subcaptionbox{}
    {\includegraphics[height=2.5cm]{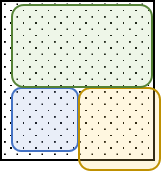}}
    \subcaptionbox{}
    {\includegraphics[height=2.5cm]{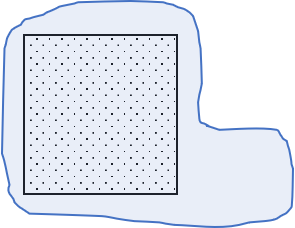}}
    \subcaptionbox{}
    {\includegraphics[height=2.5cm]{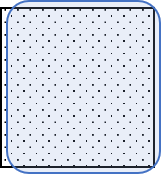}}\\
    \caption{Illustration of three overlapping relationships between a segmented object $S_j$ (colored region) and a reference object $O_i$ (dotted region). (a) Over-segmentation, (b) Under-segmentation, and (c) Matching.}
    \label{Fig.Match}
\end{figure}

In order to compare the geometric quality of a segmented object $S_j$ on the prediction map $P$ and a reference object $O_i$ on the GT map $L$, it is essential to first discriminate if they are representing the same physical object. If $S_j$ and $O_i$ are overlapped, there are three possible overlapping relationships between them, as illustrated in Fig.~\ref{Fig.Match}. Therefore, for each $O_i~(i=1,2,3,\cdots,n)$ and $S_j~(j=1,2,3,\cdots,n')$, their matching relationship $M(O_i, S_j)$ is calculated based on the over-segmentation error ($E_{os}$) and under-segmentation error ($E_{us}$)~\cite{persello2009novel}:
    \begin{equation}
        M(O_i, S_j) = \left\{
            \begin{array}{lr}
             0, & E_{os}(O_i, S_j)>T \thinspace || \thinspace E_{us}(O_i, S_j)>T\\
             1, & E_{os}(O_i, S_j)\leqslant T \thinspace \& \thinspace E_{us}(O_i, S_j)\leqslant T
             \end{array}\\
        \right.
    \end{equation}
    \begin{equation}
        E_{os}(O_i, S_j)=1-\frac{|S_j \cap O_i|}{|O_i|}, \quad
        E_{us}(O_i, S_j)=1-\frac{|S_j \cap O_i|}{|S_j|},
    \end{equation}
where $T$ is a threshold value (empirically set to 0.3). The matching rate ($MR$) of $P$ is the numeric ratio between the matched objects in $L$ and all the ${O_i}$ in $L$:
    \begin{equation}
        MR=\frac{\sum_{i,j}M(O_i, S_j)}{N_{O_i}}.
    \end{equation}
    
\begin{figure}[t]
\centering
    \subcaptionbox{}
    {\includegraphics[height=2.5cm]{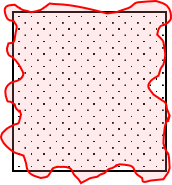}}
    \subcaptionbox{}
    {\includegraphics[height=2cm]{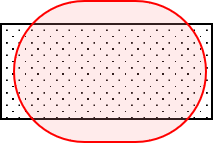}}\\
    \caption{Examples of the reference object $O_i$ (dotted region) and its matched segmented object $M_i$ (colored region) that have: (a) high curvature error ($E_{curv}$), and (b) high shape error ($E_{shape}$).}
    \label{Fig.Error}
\end{figure}

After finding the matched item $M_i$ in $P$ for $O_i$, two geometric measurements are further calculated to measure the differences between $M_i$ and $O_i$. First, $E_{curv}$ is introduced to measure the differences in object boundaries. It is calculated as:
    \begin{equation}
        E_{curv}(O_i, M_i)=||f_c(M_i)-f_c(O_i)||,
    \end{equation}
where $f_c$ denotes the contour curvature function \cite{gonzalez2002digital}. Since $O_i$ is human-annotated, $f_c(O_i)$ is usually small. A large $E_{curv}(O_i, M_i)$ indicates that the boundary of $f_c(M_i)$ is uneven. The second measurement $E_{shape}$ is introduced to assess the difference in shape, calculated as:
    \begin{equation}
        E_{shape}(O_i, M_i)=||f_s(M_i)-f_s(O_i)||, \quad f_s(M_i)=\frac{4\pi |M_i|}{{p_{M_i}}^2},
    \end{equation}
where $p_{M_i}$ is the perimeter of $M_i$. The value of $f_s(M_i)$ is 1 for a circle and $pi/4$ for a square \cite{lizarazo2014accuracy,gonzalez2002digital}. Two examples of the curvature and shape errors are illustrated in Fig.~\ref{Fig.Error}.

\section{Experimental Results}\label{sc5}
This section presents the experimental results obtained on the two VHR building datasets. First, we present the ablation study to quantitatively evaluate the improvements brought by the proposed method. Then the effects of the shape regularizer (SR) and the shape discriminator (SD) are analyzed in greater detail on some significant sample areas. Finally, the proposed ASLNet is compared with several state-of-the-art CNN models for building extraction.

\subsection{Ablation Study}\label{sc5.ablation}

\noindent \textbf{Influence of Hyper-Parameters.} The hyper-parameters $\alpha$ and $\beta$ in Eq.~\eqref{Fml.LossSeg} balance $\mathcal{L}_{pix}$ and $\mathcal{L}_{shape}$. To find which set of hyper-parameters leads to the best performance, we conduct an experiment on the Inria dataset. We set the value of one of the parameters to 1 and change the other one. The mIoU obtained with different hyper-parameter values are reported in Table \ref{Table.param}. We find that setting $\mathcal{L}_{pix}$ as the primary loss (i.e., $\alpha > \beta$) leads to higher accuracy. The ASLNet obtains the best accuracy when $\alpha = 5, \beta = 1$. Therefore, these hyper-parameters are fixed in adversarial training of the ASLNet in all the experiments.

\begin{table}
    \centering
    \begin{tabular}{c|ccccc}
        \toprule
         Hyper-parameter & 1 & 3 & 5 & 10 \\
         \hline
         $\alpha (\beta = 1)$ & 77.56 & 78.58 & \textbf{79.30} & 78.82\\
         $\beta (\alpha = 1)$ & 77.56 & 76.00 & 75.21 & 65.97 \\
        \bottomrule
    \end{tabular}
    \caption{The mIoU under different hyper-parameters tested on the Inria dataset.}
    \label{Table.param}
\end{table}

\noindent \textbf{Choice of Adversarial Losses.} There are a variety of loss functions available for the adversarial training. We test some of the commonly used losses for our task, including:
1) \textit{BCE loss}. It is calculated between the outputs of the discriminator and the domain labels (i.e., whether inputs to the discriminator are predictions or GT maps);
2) \textit{Feature Matching (FM) loss}~\cite{salimans2016improved}. It is an auxiliary loss commonly used to stabilize the training of GANs. It matches the moments of the activation on an intermediate layer of the discriminator;
3) \textit{Perceptual loss} \cite{johnson2016perceptual}. It calculates the distance between features extracted from generated and GT images using a pretrained network;
4) \textit{Multi-scale L$_1$ loss} \cite{pan2019building}. It calculates the \textit{L$_1$} distance of features in the discriminator extracted from the prediction and GT maps;
and 5) \textit{MSE loss}. It is calculated as in Eq.~\eqref{Fml.LossG}.

The obtained accuracy is reported in Table \ref{Table.AdvLoss}. The BCE loss (either w/ or w/o auxiliary loss) causes training instability and leads to unsatisfactory results, as it encourages the segmentation network to generate fake predictions unrelated to the GT situations. The perceptual loss drives the segmentation network to pay more attention to the boundary of objects (instead of the shape), since the pretrained network is not sensitive to shape features. On the contrary, the multi-scale \textit{L$_1$} loss aligns only the features without considering the segmentation maps, thus the trained network fails to optimize the building boundaries. The MSE loss successfully drives the segmentation network to learn shape patterns, leading to the highest accuracy. Therefore, it is adopted as the $\mathcal{L}_{Shape}$ to train the segmentation network.

\begin{table}
    \centering
    \begin{tabular}{c|ccccc}
        \toprule
         Adversarial Loss & OA(\%) & F1(\%) & mIoU(\%) \\
         \hline
         BCE & 96.67 & 86.23 & 76.26 \\
         BCE + FM \cite{salimans2016improved} & 96.20 & 84.67 & 73.81\\
         Perceptual \cite{johnson2016perceptual} & 96.09 & 84.38 & 73.35 \\
         Multi-scale \textit{L$_1$} \cite{pan2019building} & 96.45 & 85.67 & 75.39\\
         MSE (adopted) & \textbf{97.15} & \textbf{88.27} & \textbf{79.30} \\
        \bottomrule
    \end{tabular}
    \caption{The accuracy obtained by training with different adversarial losses on the Inria dataset.}
    \label{Table.AdvLoss}
\end{table}

\begin{figure*}[!htb]
\centering
    {\includegraphics[height=0.5cm]{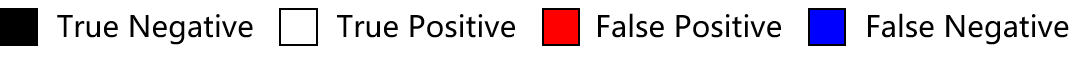}}\\
    \setlength{\tabcolsep}{1pt}
    \begin{tabular}{>{\centering\arraybackslash}m{0.4cm}>{\centering\arraybackslash}m{3.0cm}>{\centering\arraybackslash}m{3.0cm}>{\centering\arraybackslash}m{3.0cm}>{\centering\arraybackslash}m{3.0cm}>{\centering\arraybackslash}m{3.0cm}}
        (a)&
        \includegraphics[width=3.0cm]{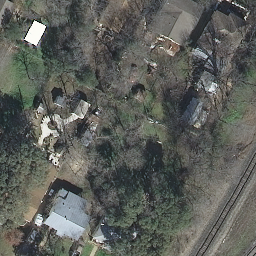} &
        \includegraphics[width=3.0cm]{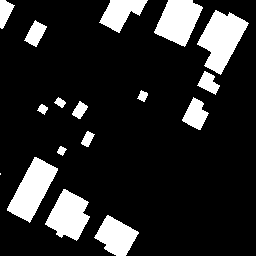} &
        \includegraphics[width=3.0cm]{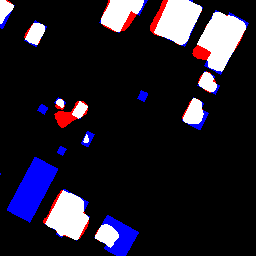} &
        \includegraphics[width=3.0cm]{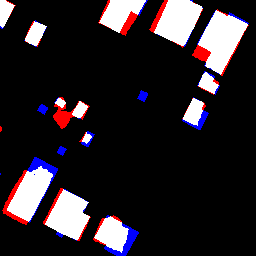} &
        \includegraphics[width=3.0cm]{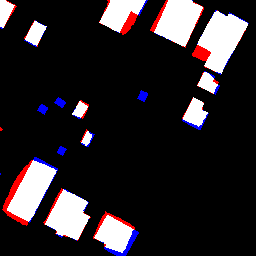}\\
        (b)&
        \includegraphics[width=3.0cm]{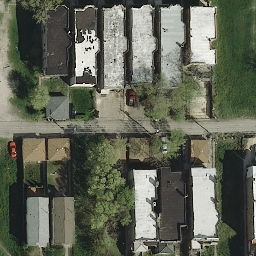} &
        \includegraphics[width=3.0cm]{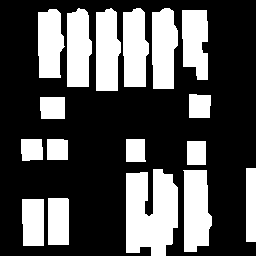} &
        \includegraphics[width=3.0cm]{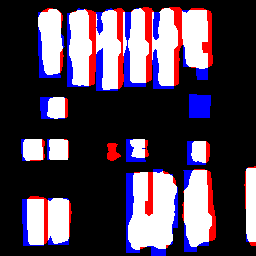} &
        \includegraphics[width=3.0cm]{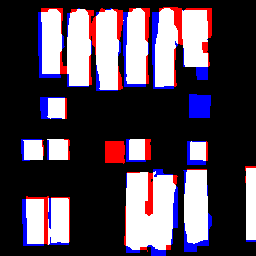} &
        \includegraphics[width=3.0cm]{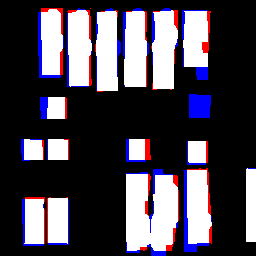}\\
        (c)&
        \includegraphics[width=3.0cm]{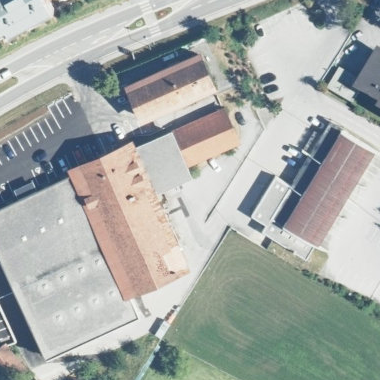} &
        \includegraphics[width=3.0cm]{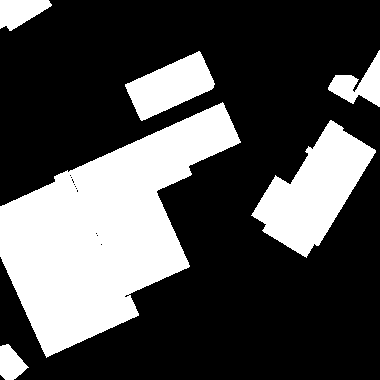} &
        \includegraphics[width=3.0cm]{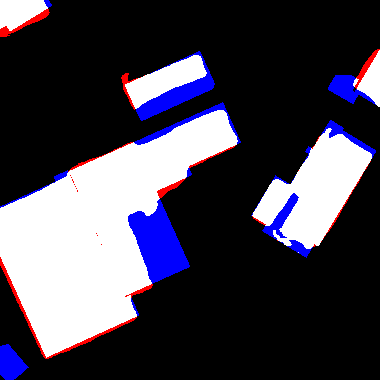} &
        \includegraphics[width=3.0cm]{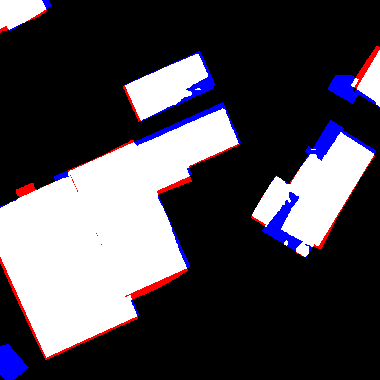} &
        \includegraphics[width=3.0cm]{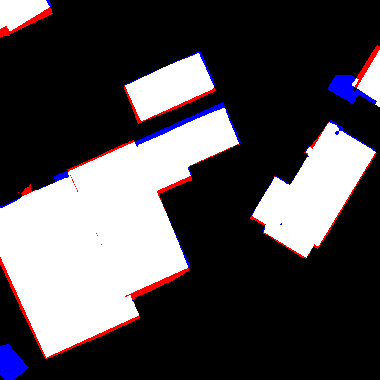}\\
        (d)&
        \includegraphics[width=3.0cm]{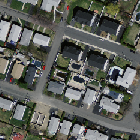} &
        \includegraphics[width=3.0cm]{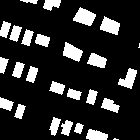} &
        \includegraphics[width=3.0cm]{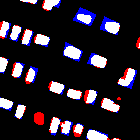} &
        \includegraphics[width=3.0cm]{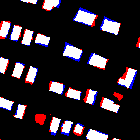} &
        \includegraphics[width=3.0cm]{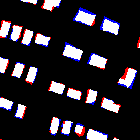}\\
        (e)&
        \includegraphics[width=3.0cm]{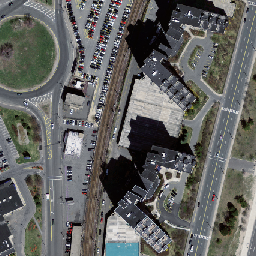} &
        \includegraphics[width=3.0cm]{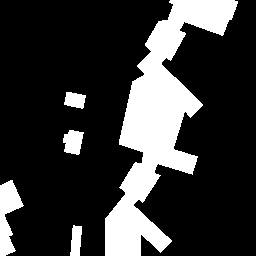} &
        \includegraphics[width=3.0cm]{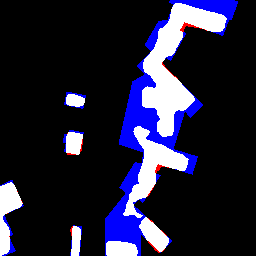} &
        \includegraphics[width=3.0cm]{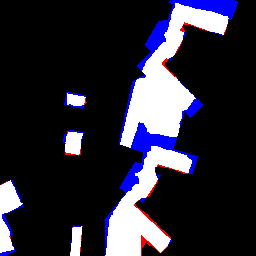} &
        \includegraphics[width=3.0cm]{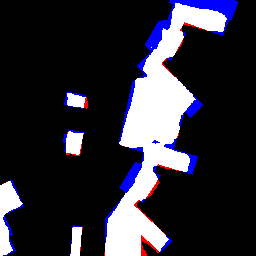}\\
        (f)&
        \includegraphics[width=3.0cm]{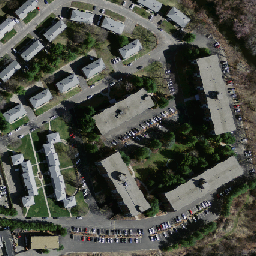} &
        \includegraphics[width=3.0cm]{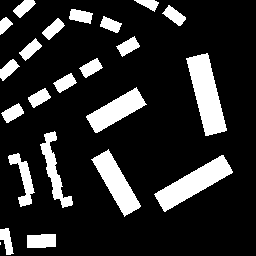} &
        \includegraphics[width=3.0cm]{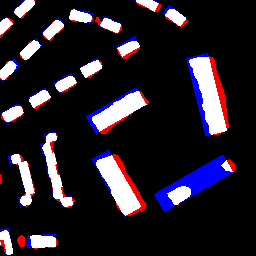} &
        \includegraphics[width=3.0cm]{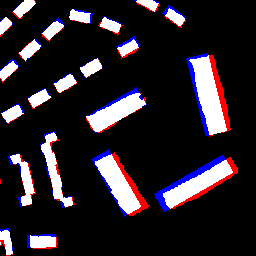} &
        \includegraphics[width=3.0cm]{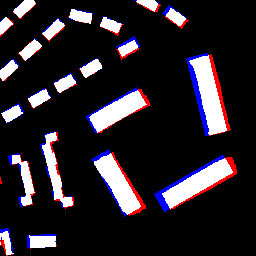}\\
        & Test Image & GT & ED-FCN & Proposed ASLNet (w/o SR) & Proposed ASLNet\\
    \end{tabular}
    \caption{Examples of segmentation results obtained by the different methods (ablation study). (a)-(c) Results selected from the Inria dataset, (d)-(f) Results selected from the Massachusetts dataset.}
    \label{Fig.AblationResults}
\end{figure*}

\begin{figure}[!htb]
\centering
    {\includegraphics[height=0.5cm]{ablation/BNsegColorBar.png}}\\
    \setlength{\tabcolsep}{1pt}
    \begin{tabular}{>{\centering\arraybackslash}m{2cm}>{\centering\arraybackslash}m{2cm}>{\centering\arraybackslash}m{2cm}>{\centering\arraybackslash}m{2cm}}
        \includegraphics[width=2cm]{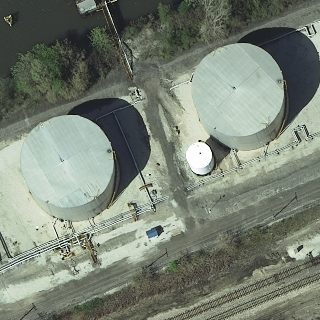} &
        \includegraphics[width=2cm]{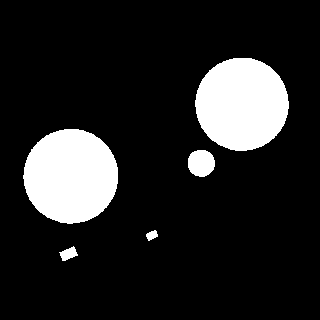} &
        \includegraphics[width=2cm]{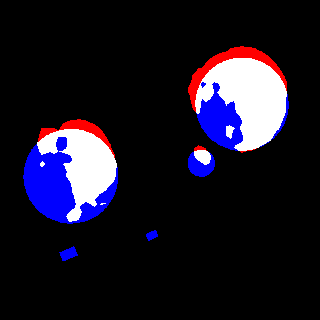} &
        \includegraphics[width=2cm]{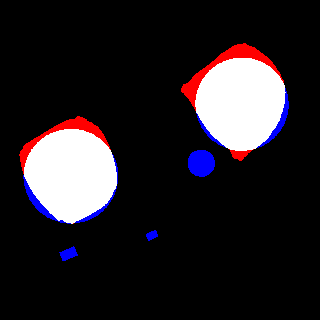}\\
        \includegraphics[width=2cm]{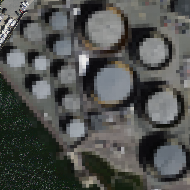} &
        \includegraphics[width=2cm]{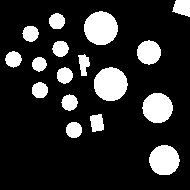} &
        \includegraphics[width=2cm]{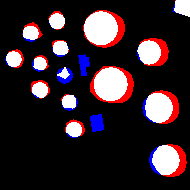} &
        \includegraphics[width=2cm]{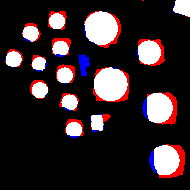}\\
        Test Image & GT & ED-FCN & Proposed ASLNet\\
    \end{tabular}
    \caption{Examples of the failure cases. The ASLNet segments rectangular items for even the round objects, given its building-shape driven training.}
    \label{Fig.FailCase}
\end{figure}

\begin{figure}[t]
\centering
        \subcaptionbox{}
        {\includegraphics[width=4.3cm]{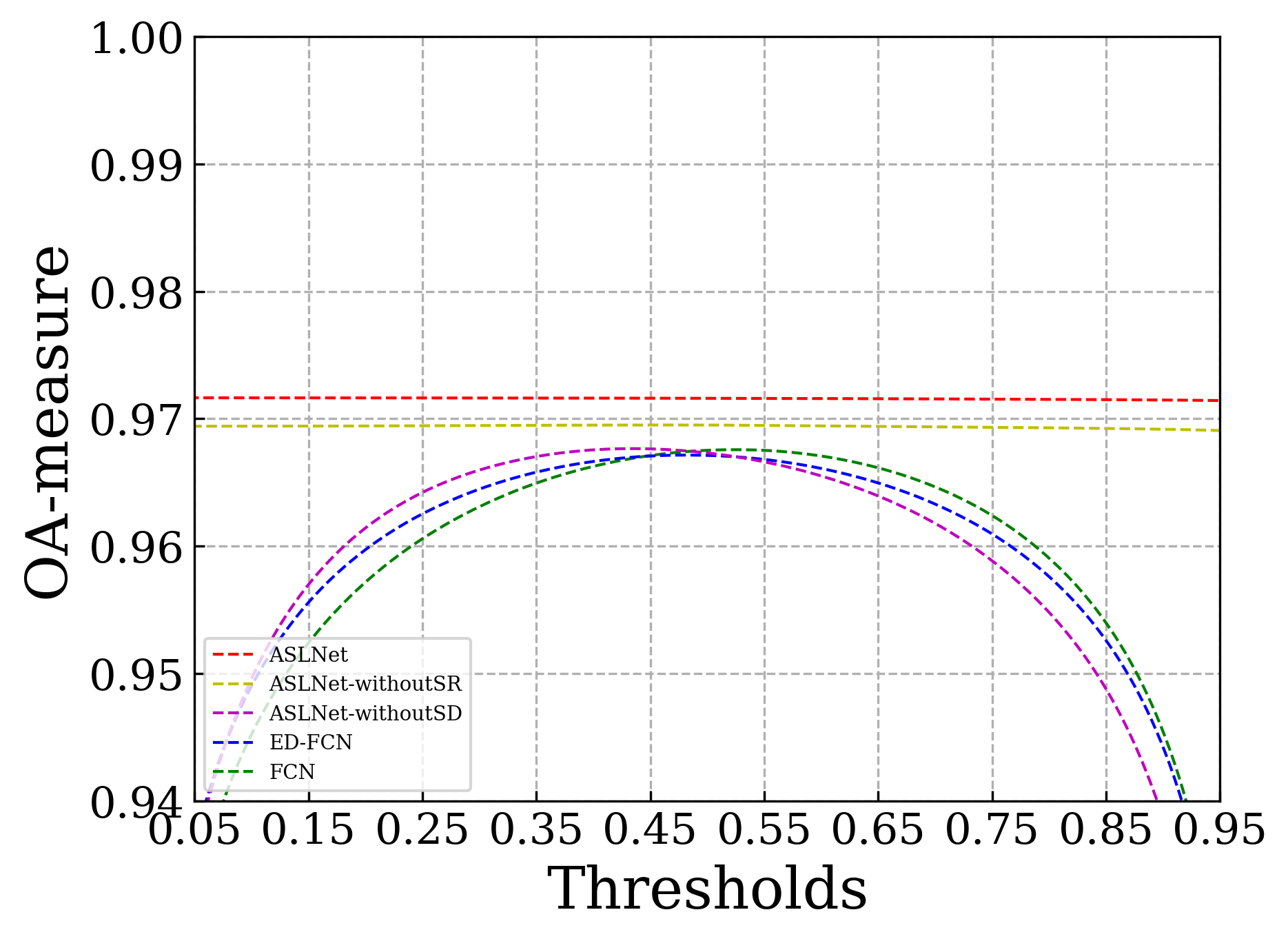}}
        \subcaptionbox{}
        {\includegraphics[width=4.3cm]{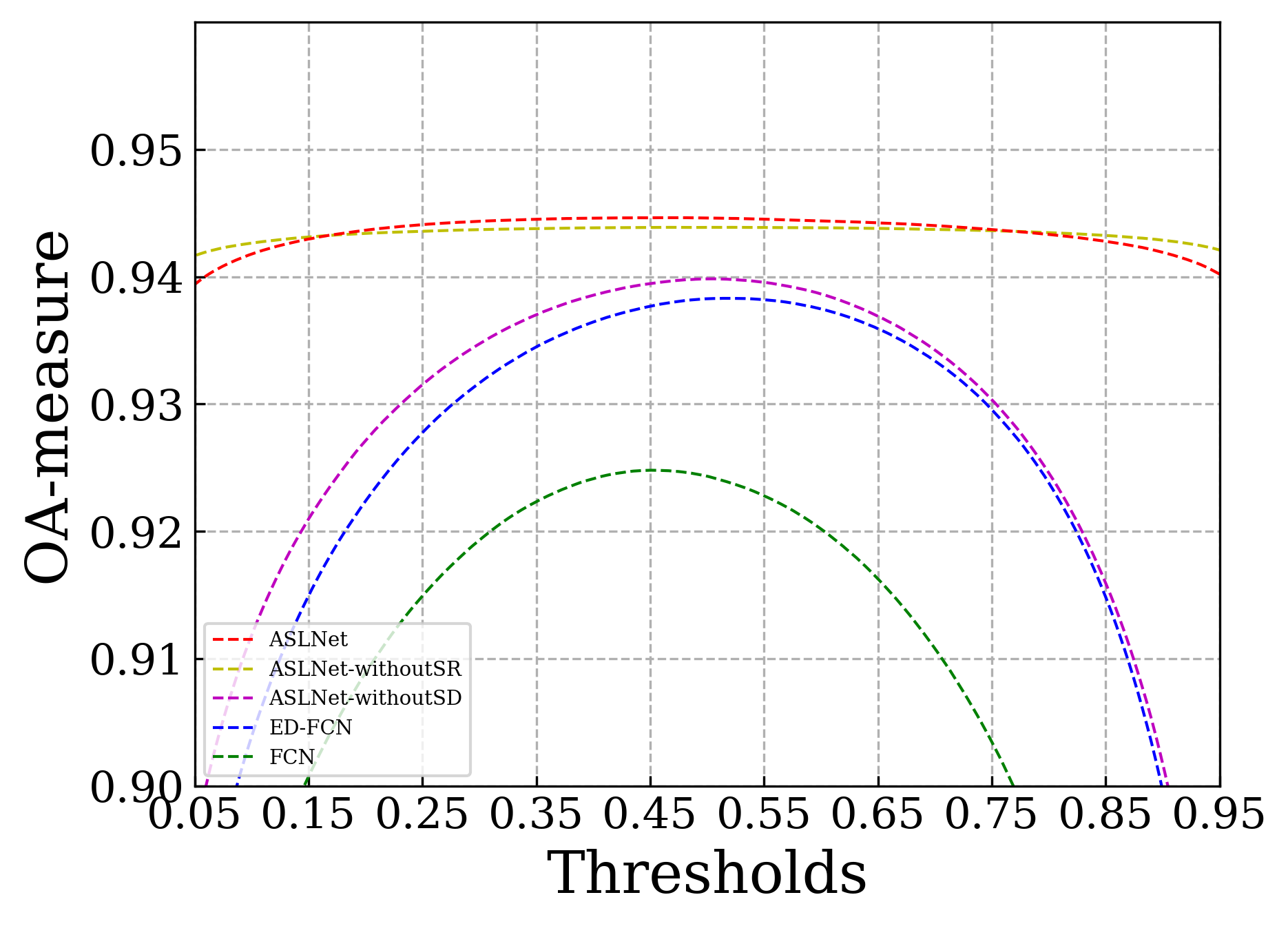}}
    \caption{Accuracy curves versus different binarization threshold of (a) Inria dataset, and (b) Massachusetts dataset.}\label{Fig.AblationCurve}
\end{figure}

\noindent \textbf{Quantitative Results.} We conduct extensive ablation studies to assess the effectiveness of the proposed ASLNet. To compare the results before and after the use of SR and SD, the original FCN \cite{long2015fcn} and the baseline method ED-FCN are also included in the comparison. The quantitative results are reported in Table~\ref{Table.Ablation}. The baseline ED-FCN outperforms the FCN in terms of mean IoU by 0.21\% and 4.87\%, respectively in the Inria and the MAS dataset, which is attributed to the concatenation of low-level features in its decoder. Since the MAS dataset has lower spatial resolution, the improvements of the ED-FCN is more noticeable. The ASLNet w/ the SR but w/o the SD has slight accuracy improvements over the ED-FCN. Meanwhile, after introducing the adversarial shape learning, the ASLNet (w/o the SR) has the mean IoU improvements of 1.56\% and 2.63\% on the two datasets. The complete ASLNet with both the SR and the SD provides improvements of 2.73\% and 3.26\% in mean IoU compared to the baseline ED-FCN. Fig.~\ref{Fig.AblationCurve} shows a comparison of the OA values of the segmented probability maps versus different binarization ($\mathcal{T}$ in Eq.~\eqref{Formula.thresholding}) thresholds. Since the ASLNet directly segments near-binary regularized results, its OA curves are close to horizontal, and are sharply above the baseline methods.

The improvements are even more significant in terms of object-based metrics. The baseline FCN encountered severe over-segmentation problems, which lead to low $MR$ values. The ED-FCN and the ASLNet (w/o the SD) slightly improve the three object-based metrics. The ASLNet (w/o the SR) has improvements of around 3\% in both $E_{curv}$ and $E_{shape}$ in the two datasets. The complete ASLNet further improves the $MR$ values of around 4\% on the two datasets.

\begin{table*}[t]
    \centering
    \caption{Results of the ablation study on the two considered data sets.}
    \resizebox{1\linewidth}{!}{%
        \begin{tabular}{c|r|c|c|ccccc|ccc}
        \toprule
            \multirow{2}*{Dataset} & \multirow{2}*{Method} & \multicolumn{2}{c|}{Components} & \multicolumn{5}{c|}{Pixel-based Metrics} & \multicolumn{3}{c}{Object-based Metrics}\\
            \cline{3-12}
            &  & SR & SD & OA(\%) & P(\%) & R(\%) & F1(\%) & mIoU(\%) & $MR$(\%) & $E_{curv}$ & $E_{shape}$\\
            \hline
            \multirow{4}*{Inria} & FCN \cite{long2015fcn} & & & 96.72 & 89.41 & 83.78 & 86.33 & 76.36 & 55.37 & 7.66 & 6.63\\
            & ED-FCN &  & &  96.69 & 87.87 & 85.29 & 86.46 & 76.57 & 60.38 & 7.26 & 6.29\\
            & Proposed ASLNet (w/o SD) & $\surd$ & & 96.71 & 87.18 & 86.77 & 86.82 & 77.06 & 57.36 & 7.42 & 6.21\\
            & Proposed ASLNet (w/o SR) & & $\surd$ & 96.94 & 88.98 & 86.32 & 87.50 & 78.13 & 60.36 & 3.86 & 4.36\\
            & Proposed ASLNet & $\surd$ & $\surd$ &\textbf{97.15} & \textbf{90.00} & \textbf{86.85} & \textbf{88.27} & \textbf{79.30} & \textbf{64.46} & \textbf{3.53} & \textbf{3.66} \\
            \hline
            \multirow{4}*{MAS} & FCN \cite{long2015fcn} & & & 92.39 & 78.46 & 78.73 & 78.56 & 64.82 & 26.87 & 11.56 & 7.79  \\
            & ED-FCN & & & 93.81 & 84.83 & 79.57 & 82.09 & 69.69 & 53.62 & 8.78 & 7.45\\
            & Proposed ASLNet (w/o SD) & $\surd$ & & 93.95 & 85.47 & 79.45 & 82.31 & 70.03 & 55.04 & 8.69 & 7.11\\
            & Proposed ASLNet (w/o SR) & & $\surd$ & 94.38 & 85.70 & 81.17 & 83.91 & 72.32 & 62.39 & 7.36 & 4.30\\
            & Proposed ASLNet & $\surd$ & $\surd$ & \textbf{94.51} & \textbf{85.92} & \textbf{82.83} & \textbf{84.32} & \textbf{72.95} & \textbf{67.28} & \textbf{7.19} & \textbf{4.01}\\
        \bottomrule
        \end{tabular} }
    \label{Table.Ablation}
\end{table*}

\noindent \textbf{Qualitative Results.} Fig.~\ref{Fig.AblationResults} shows the results of the ablation study on several sample areas. The segmentation results of the ED-FCN are generally round-edged. However, after adding the SD, the building edges became sharper and the object shapes became more rectangular. Moreover, the object shapes are modelled in a wider image range, thus the edges are more straight and some missing parts are inpainted. More specifically, Fig.~\ref{Fig.AblationResults}(a) and Fig.~\ref{Fig.AblationResults}(e) show two cases of occlusions caused by trees and shadows, respectively. Fig.~\ref{Fig.AblationResults}(c) shows a case of under-segmentation. In these cases the ASLNet has successfully recovered the complete buildings. Fig.~\ref{Fig.AblationResults}(b), (d), and (f) show several examples of the improvements in shapes. It is worth noting that the ASLNet managed to improve the segmented shape of compact small objects (e.g., houses), irregular large object (e.g., factories), and long bar-like objects (e.g., residential buildings). However, a side-effect of the ASLNet is that it fails to segment some round objects (e.g., oil tanks) that are unseen in the training set. The learned shape bias drives the ASLNet to optimize the rectangular contour of buildings. Some of examples of these cases are shown in Fig.~\ref{Fig.FailCase}. Considering the objective of the proposed method, this drawback has minor impacts. Note that the proposed shape-driven training could also be adapted to other general shapes to suit different applications.

As a conclusion of the ablation study, the modeling of shape features in the ASLNet leads to three significant benefits: 1) inpainting of the missing parts of buildings; 2) providing a joint segmentation and regularization of the building contours; 3) mitigating the under-segmentation and over-segmentation problems. These advantages are verified by both the accuracy metrics and visual observation.

\noindent \textbf{ASLNet with SOTA Backbones.} For assessing the performance of the proposed techniques, the ASLNet is designed on top of a simple ED-FCN. However, replacing the ED-FCN with more advanced segmentation networks may potentially improve its accuracy. To test this, we integrate the SR and SD modules into two well-known and widely used segmentation backbones, i.e., the DeepLabv3+ \cite{chen2018deeplabv3+} and the HRNet \cite{wang2020hrnet}. The SR module is placed at the end of each segmentation backbone, while the SD module is used in the same way as in the original ASLNet. The resulting variants of the ASLNet are referred to as the ASLNet-DL and the ASLNet-HR, respectively.

The quantitative results are reported in Table \ref{Table.DiffBackbone}. Both ASLNet-DL and ASLNet-HR obtain sharp accuracy improvements over their baselines (DeepLabv3+ and HRNet), proving that the proposed shape training method is effective on different segmentation backbones. Compared to the ASLNet, the ASLNet-HR obtains slight accuracy improvements on the two considered datasets, whereas accuracy of the ASLNet-DL is decreased on the MAS dataset. This suggests that the atrous convolutions operated on high-level features is not effective on the MAS dataset (which has a relatively lower GSD).

\begin{table}[t]
    \centering
    \caption{Results obtained using SOTA segmentation backbones.}
    \resizebox{1\linewidth}{!}{%
        \begin{tabular}{c|r|c|ccc}
        \toprule
            Dataset & Method & Backbone & OA(\%) & F1(\%) & mIoU(\%)\\
            \hline
            \multirow{4}*{Inria} & DeepLabv3+ \cite{chen2018deeplabv3+} & \multirow{2}*{DeepLabv3+} & 96.85 & 86.97 & 77.30 \\
            & ASLNet-DL (Proposed) & & \textbf{97.18} & \textbf{88.26} & \textbf{79.31} \\
            \cline{2-6}
            & HNRet \cite{wang2020hrnet} & \multirow{2}*{HNRet} & 96.90 & 87.18 & 77.68 \\
            & ASLNet-HR (Proposed) & & \textbf{97.20} & \textbf{88.40} & \textbf{79.54} \\
            \hline
            \multirow{4}*{MAS} & DeepLabv3+ \cite{chen2018deeplabv3+} & \multirow{2}*{DeepLabv3+} & 93.27 & 80.53 & 67.52 \\
            & ASLNet-DL (Proposed) & & \textbf{94.41} & \textbf{83.88} & \textbf{72.31} \\
            \cline{2-6}
            & HNRet \cite{wang2020hrnet} & \multirow{2}*{HNRet} &94.34 & 83.33 & 71.55 \\
            & ASLNet-HR (Proposed) & & \textbf{94.61} & \textbf{85.00} & \textbf{73.99}\\
        \bottomrule
        \end{tabular} }
    \label{Table.DiffBackbone}
\end{table}

\subsection{Comparative Experiments}
\begin{figure*}[!htb]
\centering
    {\includegraphics[height=0.5cm]{ablation/BNsegColorBar.png}}\\
    \setlength{\tabcolsep}{1pt}
    \begin{tabular}{>{\centering\arraybackslash}m{0.4cm}>{\centering\arraybackslash}m{2.1cm}>{\centering\arraybackslash}m{2.1cm}>{\centering\arraybackslash}m{2.1cm}>{\centering\arraybackslash}m{2.1cm}>{\centering\arraybackslash}m{2.1cm}>{\centering\arraybackslash}m{2.1cm}>{\centering\arraybackslash}m{2.1cm}>{\centering\arraybackslash}m{2.1cm}}
        (a)&
        \includegraphics[width=2.1cm]{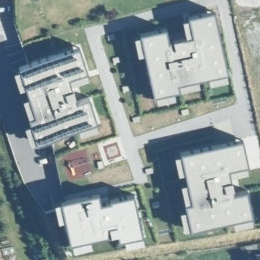} &
        \includegraphics[width=2.1cm]{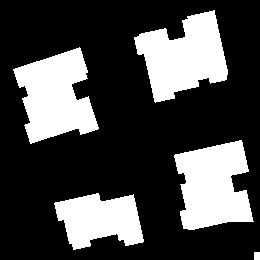} &
        \includegraphics[width=2.1cm]{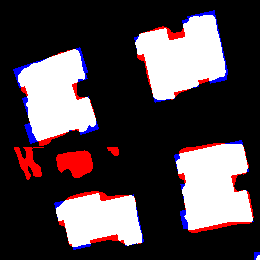} &
        \includegraphics[width=2.1cm]{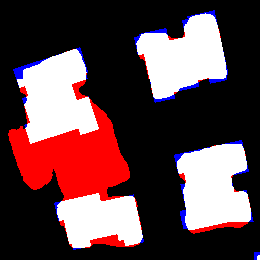} &
        \includegraphics[width=2.1cm]{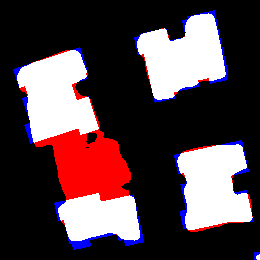} &
        \includegraphics[width=2.1cm]{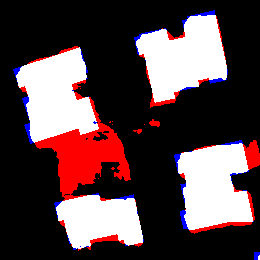} &
        \includegraphics[width=2.1cm]{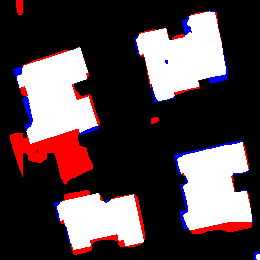} &
        \includegraphics[width=2.1cm]{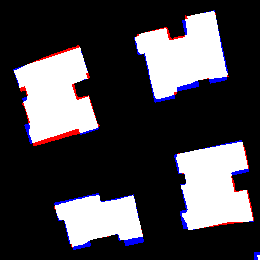}\\
        (b)&
        \includegraphics[width=2.1cm]{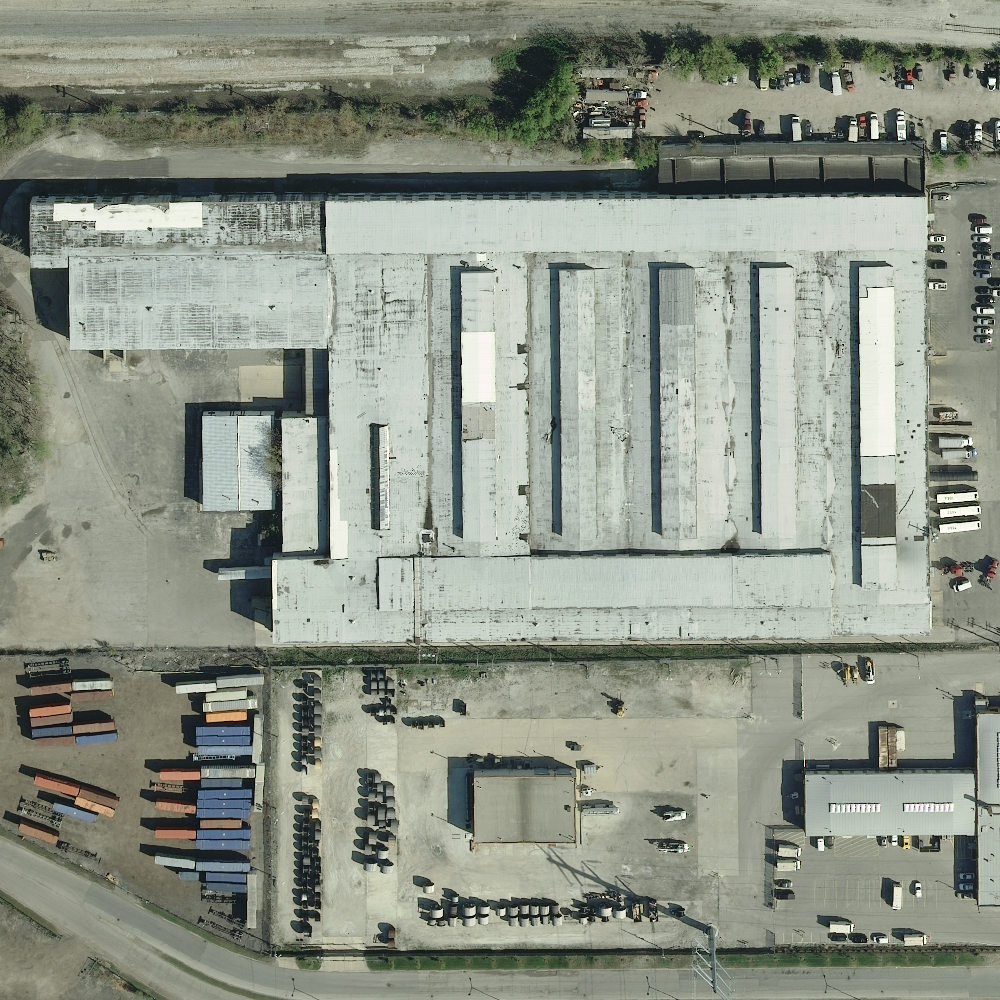} &
        \includegraphics[width=2.1cm]{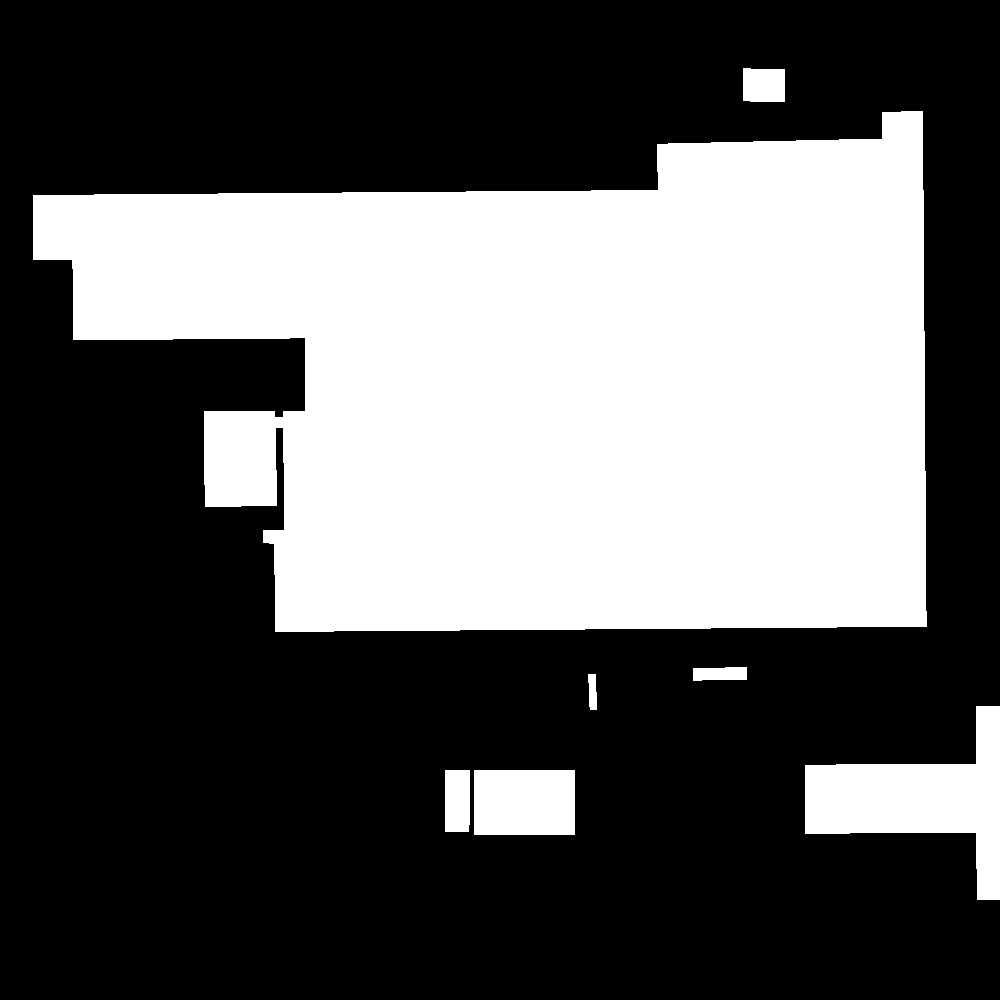} &
        \includegraphics[width=2.1cm]{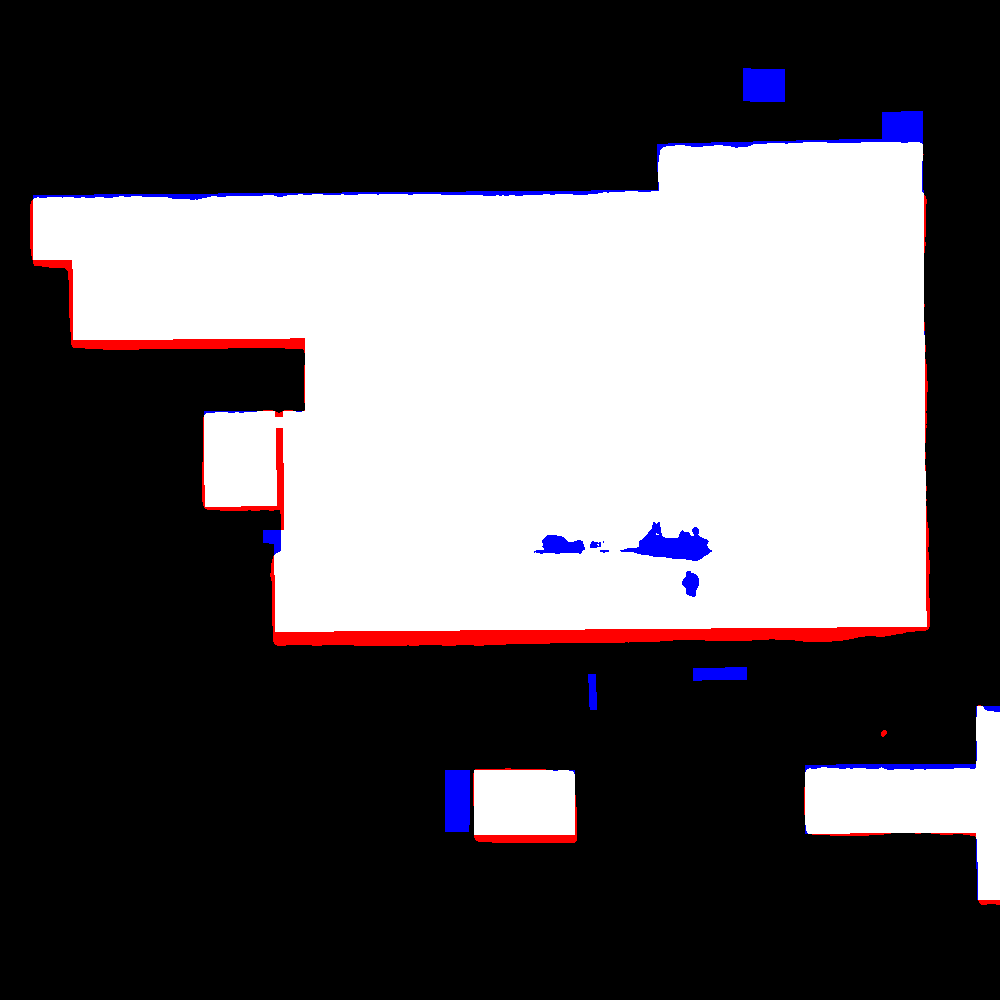} &
        \includegraphics[width=2.1cm]{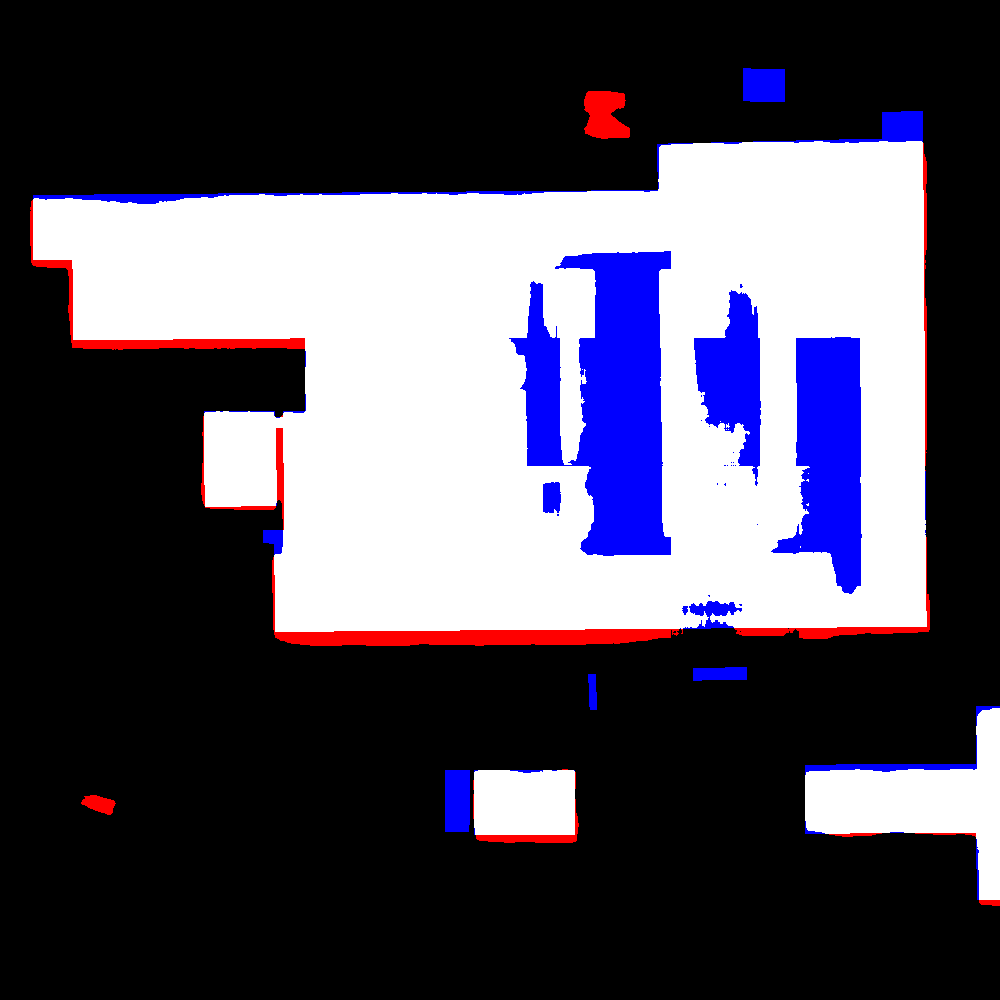} &
        \includegraphics[width=2.1cm]{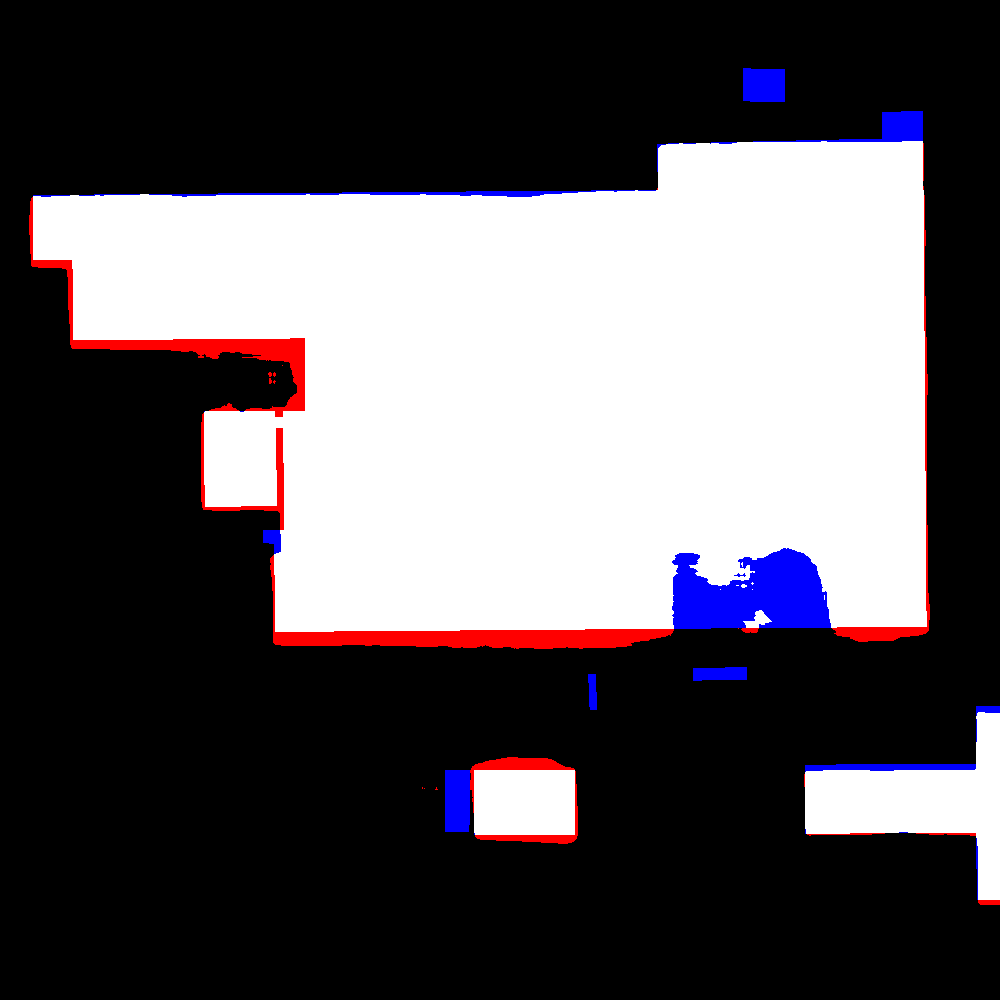} &
        \includegraphics[width=2.1cm]{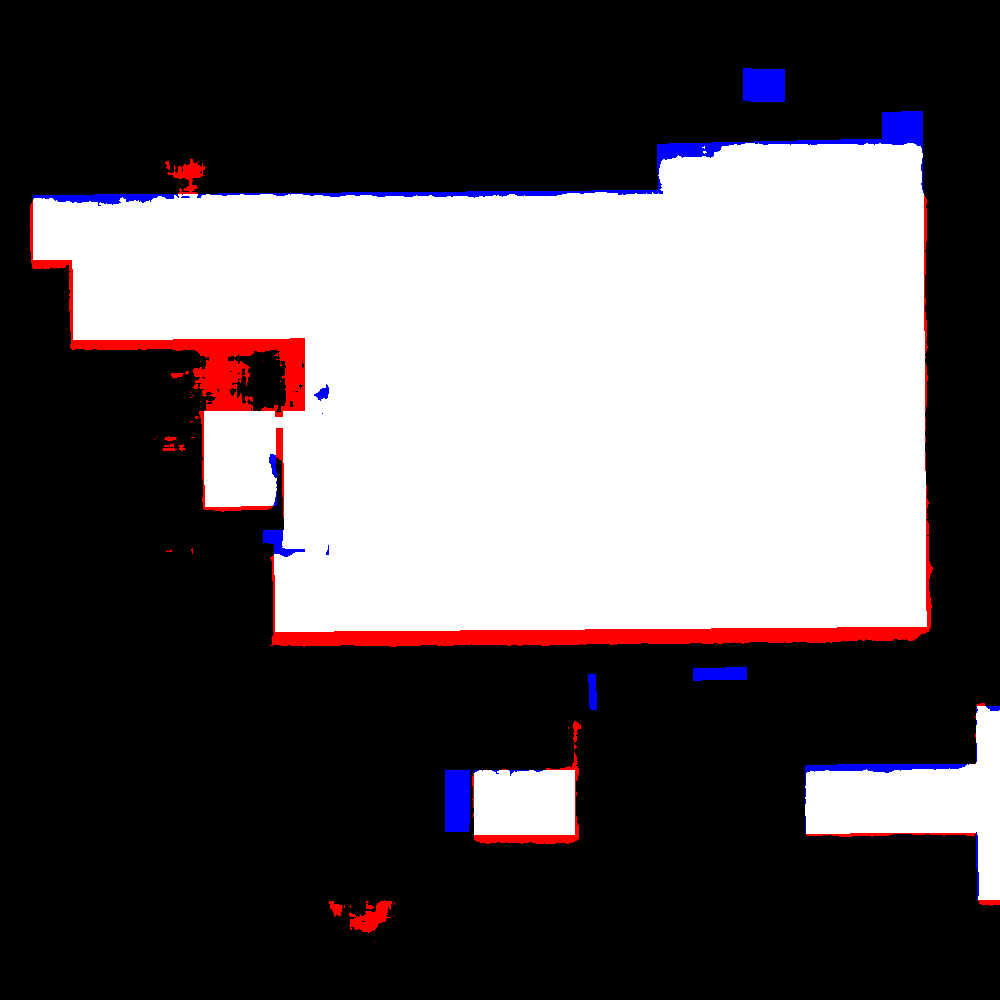} &
        \includegraphics[width=2.1cm]{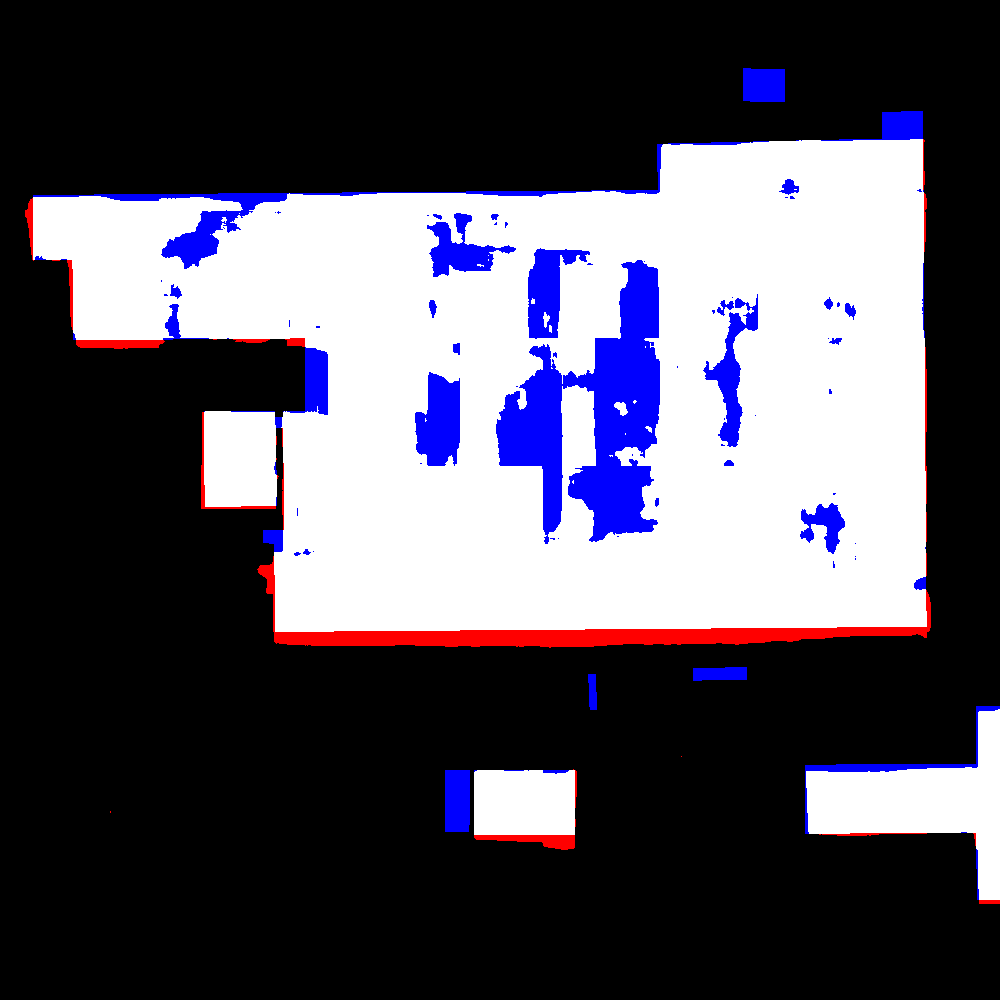} &
        \includegraphics[width=2.1cm]{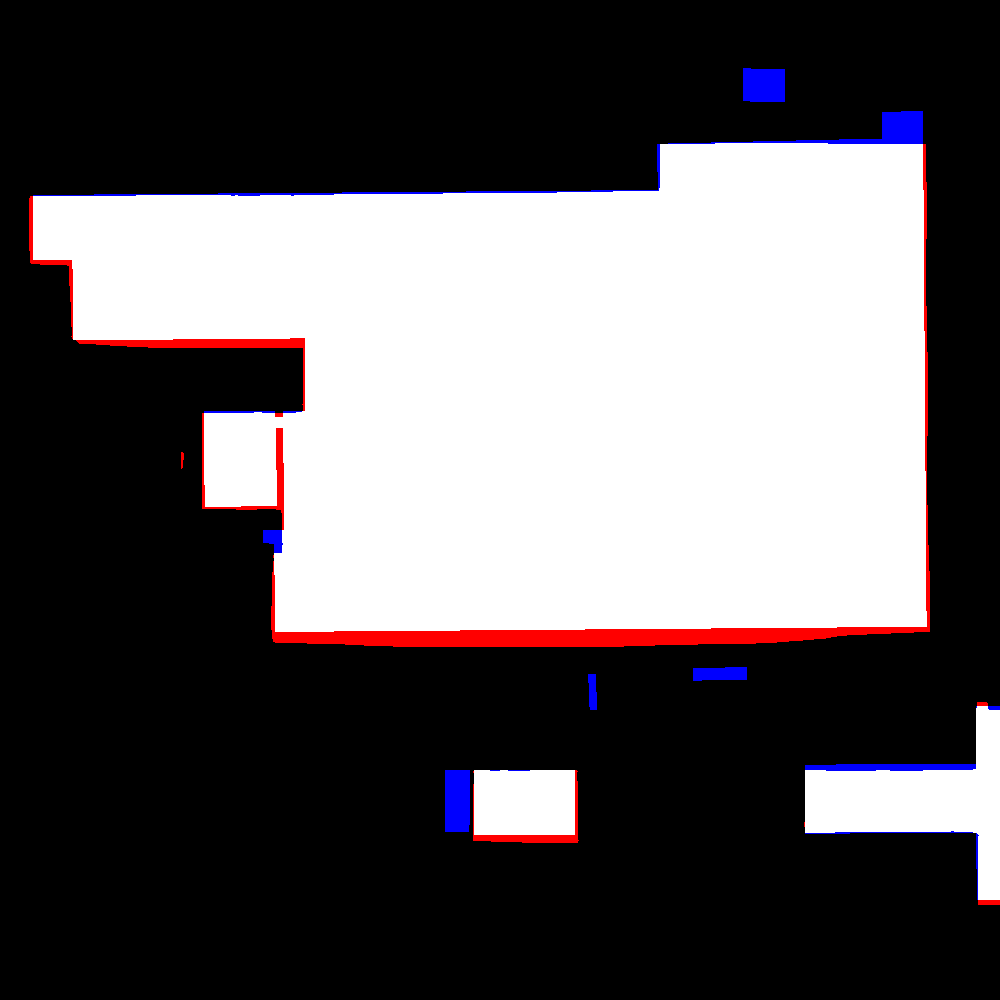}\\
        (c)&
        \includegraphics[width=2.1cm]{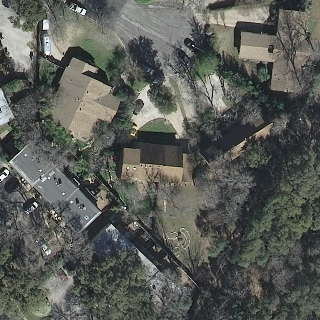} &
        \includegraphics[width=2.1cm]{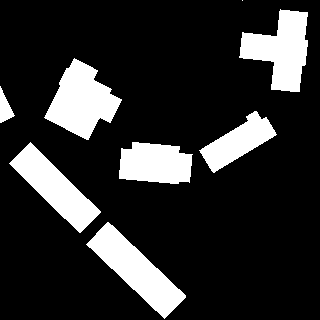} &
        \includegraphics[width=2.1cm]{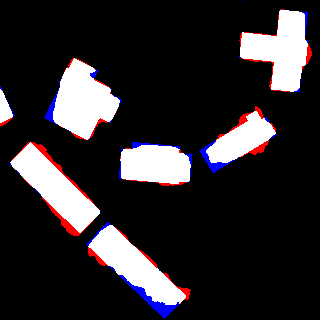} &
        \includegraphics[width=2.1cm]{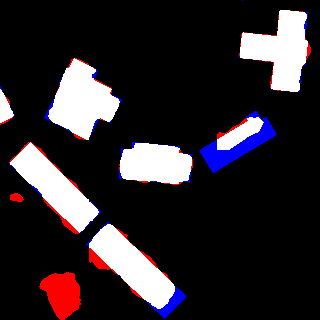} &
        \includegraphics[width=2.1cm]{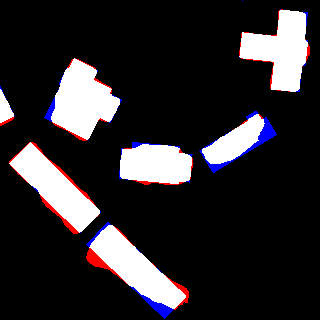} &
        \includegraphics[width=2.1cm]{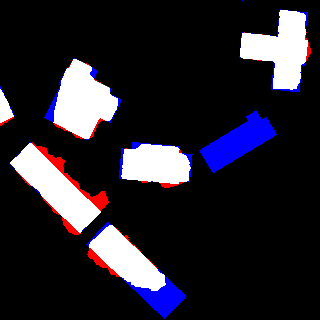} &
        \includegraphics[width=2.1cm]{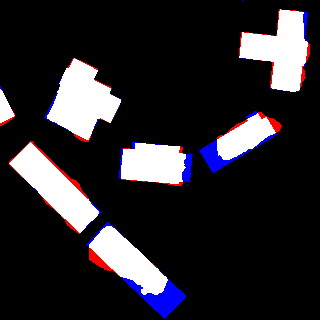} &
        \includegraphics[width=2.1cm]{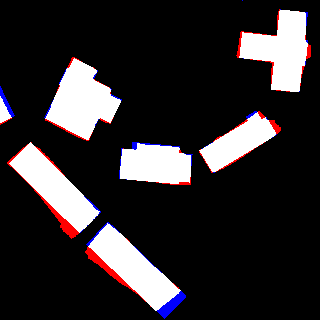}\\
        (d)&
        \includegraphics[width=2.1cm]{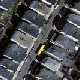} &
        \includegraphics[width=2.1cm]{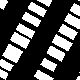} &
        \includegraphics[width=2.1cm]{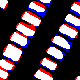} &
        \includegraphics[width=2.1cm]{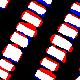} &
        \includegraphics[width=2.1cm]{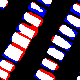} &
        \includegraphics[width=2.1cm]{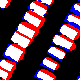} &
        \includegraphics[width=2.1cm]{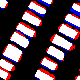} &
        \includegraphics[width=2.1cm]{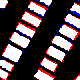}\\
        (e)&
        \includegraphics[width=2.1cm]{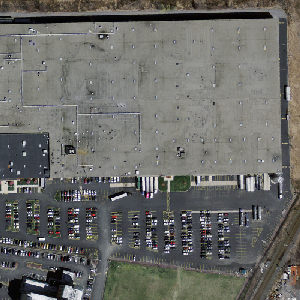} &
        \includegraphics[width=2.1cm]{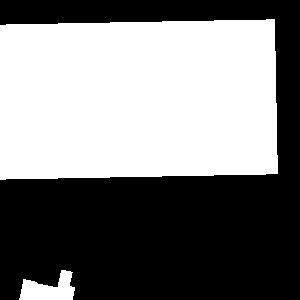} &
        \includegraphics[width=2.1cm]{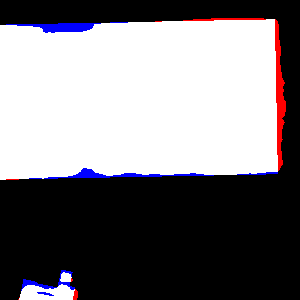} &
        \includegraphics[width=2.1cm]{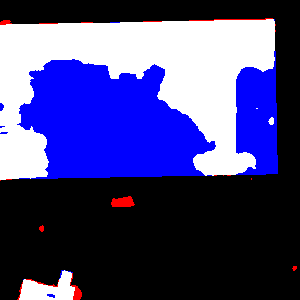} &
        \includegraphics[width=2.1cm]{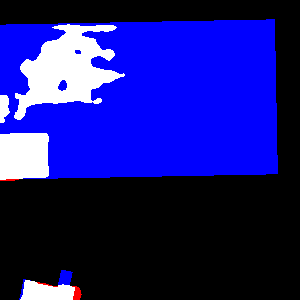} &
        \includegraphics[width=2.1cm]{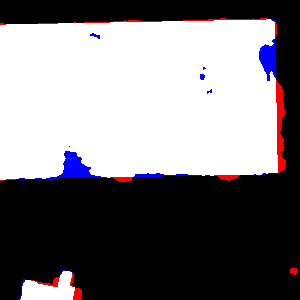} &
        \includegraphics[width=2.1cm]{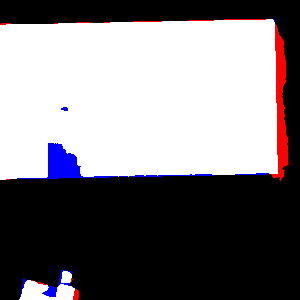} &
        \includegraphics[width=2.1cm]{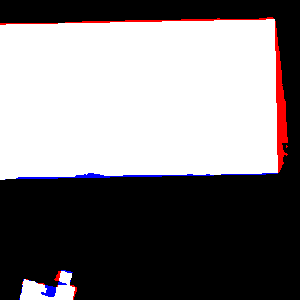}\\
        (f)&
        \includegraphics[width=2.1cm]{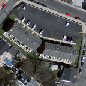} &
        \includegraphics[width=2.1cm]{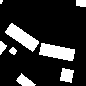} &
        \includegraphics[width=2.1cm]{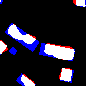} &
        \includegraphics[width=2.1cm]{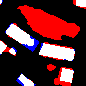} &
        \includegraphics[width=2.1cm]{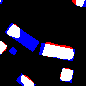} &
        \includegraphics[width=2.1cm]{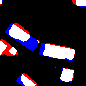} &
        \includegraphics[width=2.1cm]{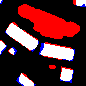} &
        \includegraphics[width=2.1cm]{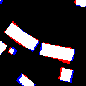}\\
        & Test Image & GT & ResUNet & cwGAN & MAPNet & GMEDN & FC-DenseNet +FRCRF & Proposed ASLNet\\
    \end{tabular}
    \caption{Examples of segmentation results obtained by the different methods (comparative experiments). (a)-(c) Results selected from the Inria dataset, (d)-(f) Results selected from the Massachusetts dataset.}
    \label{Fig.CompareResults}
\end{figure*}

\noindent \textbf{Quantitative Results.} We further compare the proposed ASLNet with several literature works to assess its effectiveness. Three classic models for the semantic segmentation are compared, including the UNet~\cite{ronneberger2015unet}, the baseline method FCN~\cite{long2015fcn} and the Deeplabv3+~\cite{chen2018deeplabv3+}. The cwGAN-gp \cite{shi2018building} that uses GAN for building extraction is also compared. Moreover, we compare the proposed method with several state-of-the-art methods for building extraction, including the ResUNet~\cite{xu2018building}, the MAPNet \cite{zhu2020map}, the GMEDN \cite{ma2020building} and the FC-DenseNet+FRCRF~\cite{li2020building} (which includes a CRF-based post-processing step). The quantitative results on the Inria dataset and the MAS dataset are reported in Table~\ref{Table.ResultInria} and Table~\ref{Table.ResultMas}, respectively.

Let us first analyze the pixel-based metrics. The ResUNet, which is a variant of UNet for the building extraction, outperforms the classic semantic segmentation models (UNet, FCN and Deeplabv3+) by a large margin on the MAS dataset. The accuracy of cw-GAN-gp is higher than that of the FCN on the MAS dataset but it is lower on the Inria dataset. on the The MAPNet obtains competitive results on the Inria dataset, whereas its performance is inferior to the ResUNet and the Deeplabv3+ on the MAS dataset. On the contrary, the GMEDN obtains better accuracy on the MAS dataset. The FCN-DenseNet+FRCRF achieves the second best accuracy on the MAS dataset. The proposed ASLNet outperforms all the compared methods in almost all the metrics (except for the precision and recall on the MAS dataset), although its baseline method (the ED-FCN) is inferior to most of them. The advantages of the ASLNet are particularly noticeable on the Inria dataset, where the ASLNet improves the mean IoU of 1.51\% with respect to the second best method. The reason for which the ASLNet has higher improvements on the Inria dataset can be attributed to the higher GSD of this dataset, where the building shape information is more discriminative.

In terms of object-based metrics, there are remarkable differences in the $MR$ values. The cw-GAN-gp and the ResUNet obtained the third best $MR$ values among the literature methods on the Inria dataset and the MAS dataset, respectively. The FCN-DenseNet+FRCRF obtained the second-best accuracy in all the object-based metrics due to its boundary-refinement CRF operations. All the other compared literature methods obtained very high $E_{curv}$ and $E_{shape}$ values. This indicates that they all suffer from irregular shapes and uneven boundaries problems. On the contrary, the proposed ASLNet shows significant advantages in terms of all these three metrics. Due to its learned shape constraints that regularize the segmented items and sharpen the building boundaries, the ASLNet exhibits great advantages in $E_{shape}$ and $E_{curv}$ in both datasets.

\noindent \textbf{Qualitative Results.} Fig.~\ref{Fig.CompareResults} shows comparisons of the segmentation results obtained by the compared methods. One can observe that the proposed ASLNet exhibits several advantages in different scenes. It is capable of accurately segmenting the individual buildings in Fig.~\ref{Fig.CompareResults}(a), the occluded houses in Fig.~\ref{Fig.CompareResults}(c) and the large-size factories/supermarkets in Fig.~\ref{Fig.CompareResults}(b) and Fig.~\ref{Fig.CompareResults}(e). When it deals with dense residential buildings as shown in Fig.~\ref{Fig.CompareResults}(d), the over-segmentation and under-segmentation errors are reduced. It also excludes some uncertain areas by considering the shape patterns (e.g., the colored opening space in Fig.~\ref{Fig.CompareResults}(a) and the parking lot in Fig.~\ref{Fig.CompareResults}(f)).

\begin{table*}[t]
    \centering
    \caption{Results of the comparative experiments on the Inria dataset.}
    \resizebox{1\linewidth}{!}{%
        \begin{tabular}{r|ccccc|ccc}
        \toprule
            \multirow{2}*{Method} & \multicolumn{5}{c|}{Pixel-based Metrics} & \multicolumn{3}{c}{Object-based Metrics} \\
            \cline{2-9}
            & OA(\%) & P(\%)  & R(\%) & F1(\%) & mIoU(\%) & $MR$(\%) & $E_{curv}$ & $E_{shape}$\\
            \hline
            UNet \cite{ronneberger2015unet} & 95.52 & 81.76 & 82.76 & 82.03 & 70.03 & 43.87 & 10.89 & 7.84 \\
            FCN \cite{long2015fcn} & 96.72 & 89.41 & 83.78 & 86.33 & 76.36 & 55.37 & 7.66 & 6.63\\
            Deeplabv3+ \cite{chen2018deeplabv3+} & 96.85 & 89.17 & 85.09 & 86.97 & 77.30 & 58.63 & 7.12 & 6.29\\
            ResUNet \cite{xu2018building} & 96.50 & 88.33 & 83.60 & 85.68 & 75.41 & 55.72 & 7.47 & 6.50\\
            cwGAN-gp \cite{shi2018building} & 96.54 & 86.43 & 85.61 & 85.94 & 75.76 & 61.51 & 7.10 & 5.47\\
            MAPNet \cite{zhu2020map} & 96.96 & 88.58 & 86.04 & 87.24 & 77.79 & 59.75 & 6.26 & 6.16\\
            GMEDN \cite{ma2020building} & 96.23 & 87.03 & 81.37 & 83.88 & 72.95 & 52.65 & 8.43 & 5.54  \\
            FC-DenseNet+FRCRF \cite{li2020building} & 96.74 & 89.55 & 83.68 & 86.36 & 76.34 & 63.43 & 4.31 & 4.13\\
            ASLNet (proposed) &\textbf{97.15} & \textbf{90.00} & \textbf{86.85} & \textbf{88.27} & \textbf{79.30} & \textbf{64.46} & \textbf{3.53} & \textbf{3.66}\\
        \bottomrule
        \end{tabular} }
    \label{Table.ResultInria}
\end{table*}

\begin{table*}[t]
    \centering
    \caption{Results of the comparative experiments on the Massachusetts dataset.}
    \resizebox{1\linewidth}{!}{%
        \begin{tabular}{r|ccccc|ccc}
        \toprule
            \multirow{2}*{Method} & \multicolumn{5}{c|}{Pixel-based Metrics} &  \multicolumn{3}{c}{Object-based Metrics}\\
            \cline{2-9}
            & OA(\%) & P(\%)  & R(\%) & F1(\%) & mIoU(\%) & $MR$(\%) & $E_{curv}$ & $E_{shape}$\\
            \hline
            UNet \cite{ronneberger2015unet} & 92.18 & 84.71 & 70.29 & 76.75 & 62.34 & 40.02 & 10.23 & 7.10 \\
            FCN \cite{long2015fcn} & 92.39 & 78.46 & 78.73 & 78.56 & 64.82 & 26.87 & 11.56 & 7.79\\
            Deeplabv3+ \cite{chen2018deeplabv3+} & 93.27 & 82.28 & 78.95 & 80.53 & 67.52 & 47.15 & 9.82 & 7.67\\
            ResUNet \cite{xu2018building} & 94.32 & 86.16 & 81.25 & 83.59 & 71.87 & 60.22 & 7.91 & 7.16\\
            cw-GAN-gp \cite{shi2018building} & 93.00 & 81.03 & 79.64 & 80.29 & 67.15 & 51.94 & 9.37 & 6.74\\
            MAPNet \cite{zhu2020map} & 93.47 & \textbf{87.88} & 72.77 & 79.50 & 66.20 & 53.70 & 8.05 & 7.63\\
            GMEDN \cite{ma2020building} & 93.29 & 84.09 & 77.49 & 80.63 & 67.61 & 51.20 & 9.20 & 7.26\\
            FC-DenseNet+FRCRF \cite{li2020building} & 94.48 & 85.28 & \textbf{83.16} & 84.18 & 72.77 & 67.21 & 7.92 & 6.66\\
            ASLNet (proposed) &\textbf{94.51} & 85.92 & 82.83 & \textbf{84.32} & \textbf{72.95} & \textbf{67.28} & \textbf{7.19} & \textbf{4.01} \\
        \bottomrule
        \end{tabular} }
    \label{Table.ResultMas}
\end{table*}

\section{Conclusions}\label{sc6}

Recent works on CNN-based building extraction exhibit severe limitations resulting in two main issues: 1) incomplete segmentation of objects due to occlusions and intra-class diversity; 2) geometric regularization of the building extraction results. To address these issues, we introduce the adversarial training strategy to learn the shape of buildings and propose an ASLNet. Specifically, we designed a SR with shape-sensitive convolutional layers (DCs and DFCs) to regularize the feature maps, as well as a SD to learn the shape constraints to guide the segmentation network. The SR and SD allow an accurate modelling of the shape information contained in the considered images. To the best of our knowledge, this is the first work that learns adversarial shape constraints for the segmentation of RSIs. To quantitatively evaluate the thematic properties of the building extraction results, we also designed three object-based metrics: the matching rate, the curvature error and the shape error.

Experimental results on two VHR building datasets show that the proposed ASLNet has obtained significant improvements over the conventional CNN models in both pixel-based metrics and object-based metrics. These improvements can be attributed to two factors. First, learning the shape priors is beneficial to inpaint the missing building parts. Second, the shape constraints force the ASLNet to produce shape-regularized results, thus the segmented objects have rectangular shape and smooth boundaries. Additionally, we observed that the ASLNet greatly reduces the over-segmentation and under-segmentation errors (proved by the higher $MR$ values). One of the limitation of the ASLNet is that it reduces its accuracy on the segmentation of objects with shape that are not rectangular (e.g., round buildings), which is due to its learned shape constraints.

The adversarial shape learning is potentially beneficial for other segmentation-related tasks with the RSIs, where the ground objects exhibit certain geometric patterns. In future studies, we will investigate to use the adversarial shape learning to model other types of object shapes in different tasks (e.g., road extraction, change detection and land-cover mapping in RSIs).

\bibliographystyle{IEEEtran}
\bibliography{refs}


\begin{IEEEbiography}[{\includegraphics[width=1in,height=1.25in,clip,keepaspectratio]{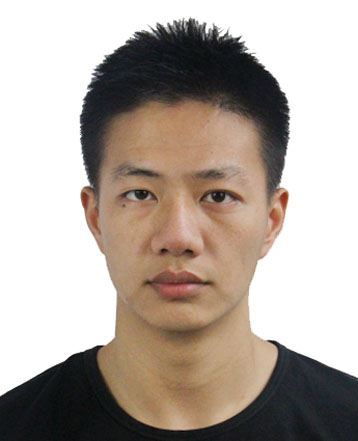}}]{Lei Ding}
received the MS’s degree in Photogrammetry and Remote Sensing from the Information Engineering University (Zhengzhou, China), and the PhD (cum laude) in Communication and Information Technologies from the University of Trento (Trento, Italy). His research interests are related to semantic segmentation, change detection and domain adaptation with Deep Learning techniques. He is a referee for many international journals, including IEEE TIP, TNNLS, TGRS, GRSL and JSTAR.
\end{IEEEbiography}

\begin{IEEEbiography}[{\includegraphics[width=1in,height=1.25in,clip,keepaspectratio]{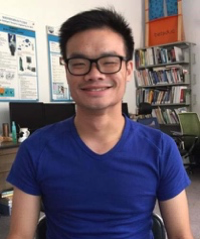}}]{Hao Tang}
is currently a Postdoctoral with Computer Vision Lab, ETH Zurich, Switzerland. 
He received the master’s degree from the School of Electronics and Computer Engineering, Peking University, China and the Ph.D. degree from Multimedia and Human Understanding Group, University of Trento, Italy.
He was a visiting scholar in the Department of Engineering Science at the University of Oxford. His research interests are deep learning, machine learning, and their applications to computer vision.
\end{IEEEbiography}

\begin{IEEEbiography}[{\includegraphics[width=1in,height=1.25in,clip,keepaspectratio]{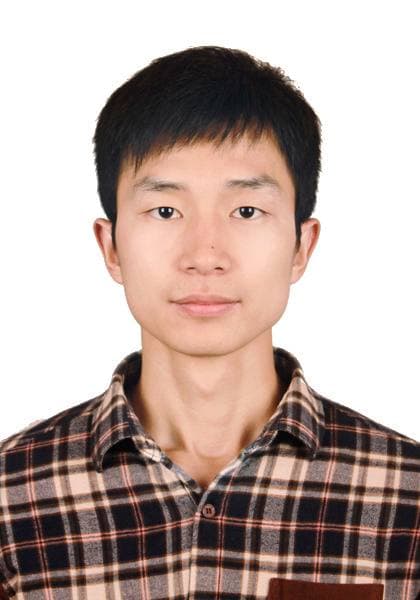}}]{Yahui Liu}
is a Ph.D. student in the Department of Information Engineering and Computer Science at the University of Trento, Italy. Before that, he received his B.S. degree and M.S. degree from Wuhan University, China, in 2015 and 2018, respectively. His major research interests are machine learning and computer vision, including unsupervised learning and image domain translation.
\end{IEEEbiography}

\begin{IEEEbiography}[{\includegraphics[width=1in,height=1.25in,clip,keepaspectratio]{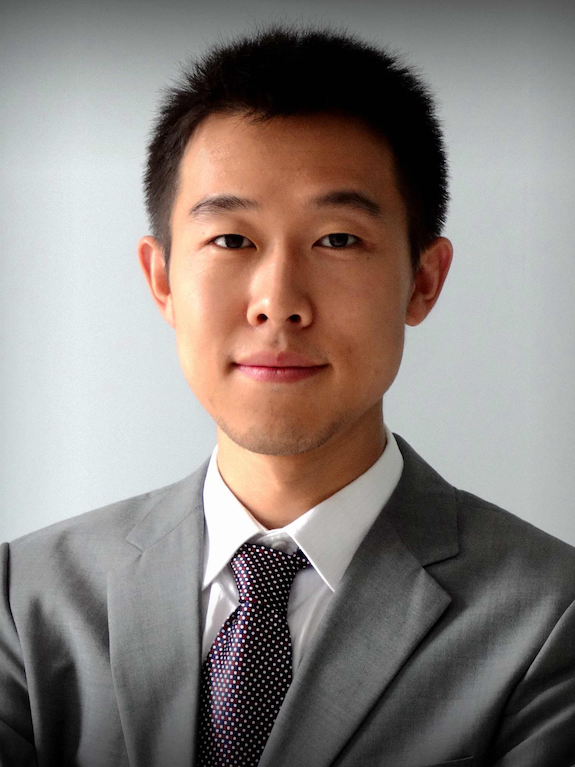}}]{Yilei Shi}(M'18) received his Diploma (Dipl.-Ing.) degree in Mechanical Engineering, his Doctorate (Dr.-Ing.) degree in Engineering from Technical University of Munich (TUM), Germany. In April and May 2019, he was a guest scientist with the department of applied mathematics and theoretical physics, University of Cambridge, United Kingdom. He is currently a senior scientist with the Chair of Remote Sensing Technology, Technical University of Munich.

His research interests include computational intelligence, fast solver and parallel computing for large-scale problems, advanced methods on SAR and InSAR processing, machine learning and deep learning for variety data sources, such as SAR, optical images, medical images and so on; PDE related numerical modeling and computing.
\end{IEEEbiography}

\begin{IEEEbiography}[{\includegraphics[width=1in,height=1.25in,clip,keepaspectratio]{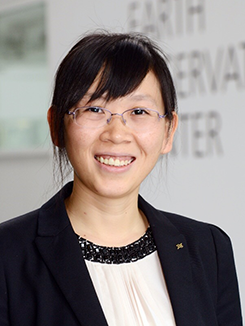}}]{Xiao Xiang Zhu}(S'10--M'12--SM'14--F'21) received the Master (M.Sc.) degree, her doctor of engineering (Dr.-Ing.) degree and her “Habilitation” in the field of signal processing from Technical University of Munich (TUM), Munich, Germany, in 2008, 2011 and 2013, respectively.
\par
She is currently the Professor for Data Science in Earth Observation (former: Signal Processing in Earth Observation) at Technical University of Munich (TUM) and the Head of the Department ``EO Data Science'' at the Remote Sensing Technology Institute, German Aerospace Center (DLR). Since 2019, Zhu is a co-coordinator of the Munich Data Science Research School (www.mu-ds.de). Since 2019 She also heads the Helmholtz Artificial Intelligence -- Research Field ``Aeronautics, Space and Transport". Since May 2020, she is the director of the international future AI lab "AI4EO -- Artificial Intelligence for Earth Observation: Reasoning, Uncertainties, Ethics and Beyond", Munich, Germany. Since October 2020, she also serves as a co-director of the Munich Data Science Institute (MDSI), TUM. Prof. Zhu was a guest scientist or visiting professor at the Italian National Research Council (CNR-IREA), Naples, Italy, Fudan University, Shanghai, China, the University of Tokyo, Tokyo, Japan and University of California, Los Angeles, United States in 2009, 2014, 2015 and 2016, respectively. She is currently a visiting AI professor at ESA's Phi-lab. Her main research interests are remote sensing and Earth observation, signal processing, machine learning and data science, with a special application focus on global urban mapping.

Dr. Zhu is a member of young academy (Junge Akademie/Junges Kolleg) at the Berlin-Brandenburg Academy of Sciences and Humanities and the German National Academy of Sciences Leopoldina and the Bavarian Academy of Sciences and Humanities. She is an associate Editor of IEEE Transactions on Geoscience and Remote Sensing and serves as the area editor responsible for special issues of IEEE Signal Processing Magazine. She is a Fellow of IEEE.
\end{IEEEbiography}

\begin{IEEEbiography}[{\includegraphics[width=1in,height=1.25in,clip,keepaspectratio]{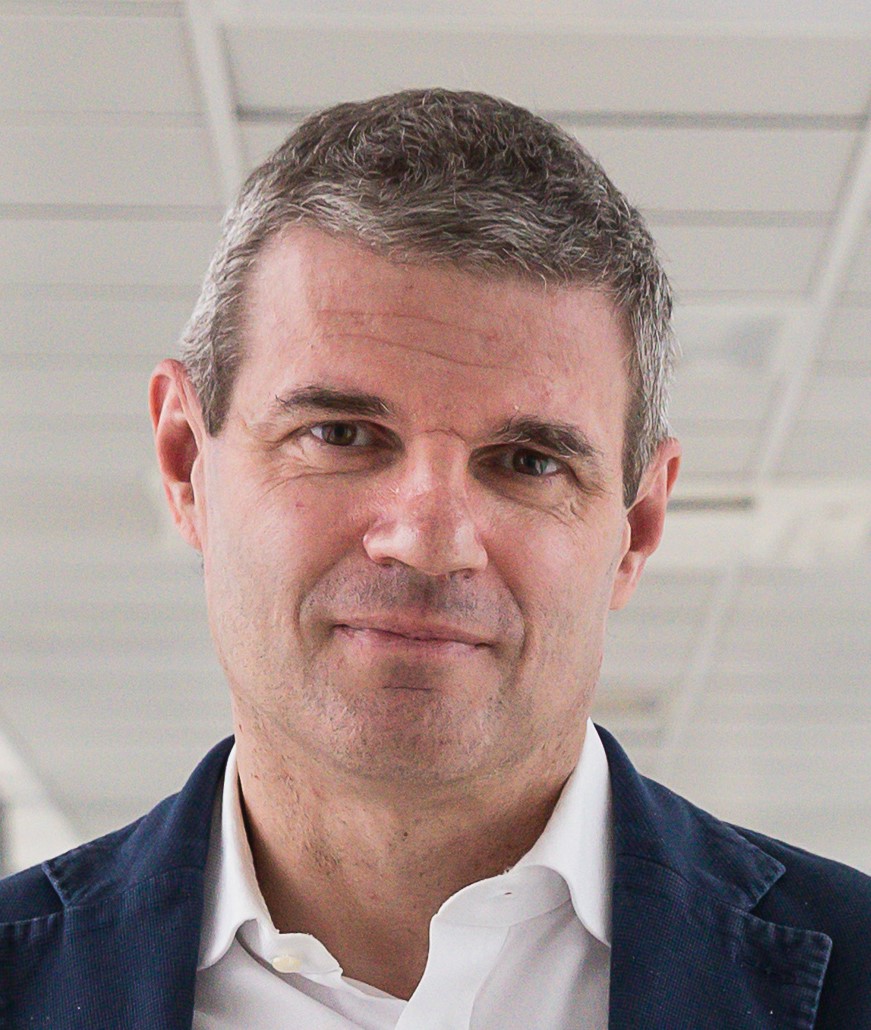}}]{Lorenzo Bruzzone}
(S'95-M'98-SM'03-F'10) received the Laurea (M.S.) degree in electronic engineering (\emph{summa cum laude}) and the Ph.D. degree in telecommunications from the University of Genoa, Italy, in 1993 and 1998, respectively. \\
He is currently a Full Professor of telecommunications at the University of Trento, Italy, where he teaches remote sensing, radar, and digital communications. Dr. Bruzzone is the founder and the director of the Remote Sensing Laboratory in the Department of Information Engineering and Computer Science, University of Trento. His current research interests are in the areas of remote sensing, radar and SAR, signal processing, machine learning and pattern recognition. He promotes and supervises research on these topics within the frameworks of many national and international projects. He is the Principal Investigator of many research projects. Among the others, he is the Principal Investigator of the \emph{Radar for icy Moon exploration} (RIME) instrument in the framework of the \emph{JUpiter ICy moons Explorer} (JUICE) mission of the European Space Agency. He is the author (or coauthor) of 215 scientific publications in referred international journals (154 in IEEE journals), more than 290 papers in conference proceedings, and 21 book chapters. He is editor/co-editor of 18 books/conference proceedings and 1 scientific book. He was invited as keynote speaker in more than 30 international conferences and workshops. Since 2009 he is a member of the Administrative Committee of the IEEE Geoscience and Remote Sensing Society (GRSS). 

Dr. Bruzzone was a Guest Co-Editor of many Special Issues of international journals. He is the co-founder of the IEEE International Workshop on the Analysis of Multi-Temporal Remote-Sensing Images (MultiTemp) series and is currently a member of the Permanent Steering Committee of this series of workshops. Since 2003 he has been the Chair of the SPIE Conference on Image and Signal Processing for Remote Sensing. He has been the founder of the IEEE Geoscience and Remote Sensing Magazine for which he has been Editor-in-Chief between 2013-2017. Currently he is an Associate Editor for the IEEE Transactions on Geoscience and Remote Sensing. He has been Distinguished Speaker of the IEEE Geoscience and Remote Sensing Society between 2012-2016. His papers are highly cited, as proven form the total number of citations (more than 27000) and the value of the h-index (78) (source: Google Scholar).
\end{IEEEbiography}

\end{document}